\documentclass[letterpaper, 10 pt, conference]{ieeeconf}  

\IEEEoverridecommandlockouts                              

\usepackage{hyperref}
\usepackage{url}
\usepackage{graphics} 
\usepackage{times} 
\usepackage{booktabs}       
\usepackage{amsfonts}       
\usepackage{nicefrac}       
\usepackage{microtype}      
\usepackage{algorithm}
\usepackage{algorithmic}
\usepackage{mathtools}
\usepackage{graphicx}
\usepackage{cases}
\usepackage{array}
\usepackage{color}
\usepackage{float}
\usepackage{amssymb}

\usepackage{wrapfig}
\usepackage{subfigure}
\usepackage{amsmath}
\usepackage{soul}

\newcommand{\define}{\triangleq}
\newtheorem{theorem}{Theorem}
\newtheorem{definition}{Definition}

\newtheorem{assumption}{Assumption}

\newtheorem{remark}{Remark}

\makeatletter
\def\UrlAlphabet{%
      \do\a\do\b\do\c\do\d\do\e\do\f\do\g\do\h\do\i\do\j%
      \do\k\do\l\do\m\do\n\do\o\do\p\do\q\do\r\do\s\do\t%
      \do\u\do\v\do\w\do\x\do\y\do\z\do\A\do\B\do\C\do\D%
      \do\E\do\F\do\G\do\H\do\I\do\J\do\K\do\L\do\M\do\N%
      \do\O\do\P\do\Q\do\R\do\S\do\T\do\U\do\V\do\W\do\X%
      \do\Y\do\Z}
\def\UrlDigits{\do\1\do\2\do\3\do\4\do\5\do\6\do\7\do\8\do\9\do\0}
\g@addto@macro{\UrlBreaks}{\UrlOrds}
\g@addto@macro{\UrlBreaks}{\UrlAlphabet}
\g@addto@macro{\UrlBreaks}{\UrlDigits}
\makeatother

\title{\LARGE \bf
Actor-Critic Reinforcement Learning for Control \\ with Stability Guarantee 
}

\author{Minghao Han$^{1}$,  Lixian Zhang$^{1}$, Jun Wang$^{2}$, and Wei Pan$^{3}$
\thanks{$^{1}$Minghao Han and Lixian Zhang are with the Department of Control Science and Engineering, Harbin Institute of Technology, China. {\tt\small mhhan@hit.edu.cn,lixianzhang@hit.edu.cn.}}%
\thanks{$^{2}$Jun Wang is with the Department of Computer Science, University College London, UK. \texttt{\small jun.wang@cs.ucl.ac.uk}.}
\thanks{$^{3}$Wei Pan is with the Department of Cognitive Robotics, Delft University of Technology, Netherlands. \texttt{\small wei.pan@tudelf.nl}.}%
}

\begin{document}

\maketitle
\thispagestyle{empty}
\pagestyle{empty}

\begin{abstract}
  
Reinforcement Learning (RL) and its integration with deep learning have achieved impressive performance in various robotic control tasks, ranging from motion planning and navigation to end-to-end visual manipulation. However, stability is not guaranteed in model-free RL by solely using data. From a control-theoretic perspective, stability is the most important property for any control system, since it is closely related to safety, robustness, and reliability of robotic systems. In this paper, we propose an actor-critic RL framework for control which can guarantee closed-loop stability by employing the classic Lyapunov's method in control theory. First of all, a data-based stability theorem is proposed for stochastic nonlinear systems modeled by Markov decision process. Then we show that the stability condition could be exploited as the critic in the actor-critic RL to learn a controller/policy. At last, the effectiveness of our approach is evaluated on several well-known 3-dimensional robot control tasks and a synthetic biology gene network tracking task in three different popular physics simulation platforms. As an empirical evaluation on the advantage of stability, we show that the learned policies can enable the systems to recover to the equilibrium or way-points when interfered by uncertainties such as system parametric variations and external disturbances to a certain extent.

\end{abstract}
\section{INTRODUCTION}
Reinforcement learning is promising to highly nonlinear control robotic systems with large state and action space \cite{kober2013reinforcement}. Until recently, significant progress has been made by combining advances in deep learning with reinforcement learning. Impressive results are obtained in a series of high-dimensional robotic control tasks where sophisticated and hard-to-engineer behaviors can be achieved~\cite{peters2008reinforcement,lockel2020probabilistic,zhu2017target,gu2017deep}. However, the performance of an RL agent is by large evaluated through trial-and-error and RL could hardly provide any guarantee for the reliability of the learned control policy.

Given a control system, regardless of which controller design method is used, the first and most important property of a system needs to be guaranteed is stability, because an unstable control system is typically useless and potentially dangerous~\cite{slotine1991applied}. A stable system is guaranteed to converge to the equilibrium or reference signal and it could recover to these targets even in the presence of parametric uncertainties and disturbances \cite{vidyasagar2002nonlinear}. Thus stability is closely related to the robustness, safety, and reliability of the robotic systems.

The most useful and general approach for studying the stability of robotic systems is Lyapunov's method \cite{lyapunov1892general}, which is dominant in control engineering~\cite{aastrom1989adaptive,mayne2000constrained}. In Lyapunov's method, a scalar ``energy-like" function called Lyapunov function $L$ is constructed to analyze the stability of the system. The controller is designed to ensure that the difference of Lyapunov function along the state trajectory is negative definite for all time instants so that the state goes in the direction of decreasing the value of Lyapunov function and eventually converges to the equilibrium \cite{corless1981continuous,thowsen1983uniform}. In learning methods, as the dynamic model is unknown, the ``energy decreasing'' condition has to be verified by trying out \emph{all} possible consecutive data pairs in the state space, i.e., to verify infinite inequalities $L_{t+1} - L_t < 0$. Obviously, the ``infinity'' requirement is impossible thus making the direct exploitation of Lyapunov's method impossible.

In this paper, we propose a data-based stability theorem and an actor-critic reinforcement learning algorithm to jointly learn the controller/policy and a Lyapunov critic function both of which are parameterized by deep neural networks, with a focus on stabilization and tracking tasks in robotic systems. The contribution of our paper can be summarized as follows: \textcolor{black}{1) a novel data-based stability theorem where only one inequality on the expected value over the state space needs to be evaluated; 2) a sample approximation of the stability condition proposed above is exploited to derive an actor-critic algorithm to search for a controller with asymptotic stability guarantee (in the number of data points);} 3) we show through experiments that the learned controller could stabilize the systems when interfered by uncertainties such as unseen disturbances and system parametric variations of a certain extent. In our experiment, we show that the stability guaranteed controller is more capable of handling uncertainties compared to those without such guarantees in nonlinear control problems including classic CartPole stabilization tasks, control of 3D legged robots and manipulator, and reference tracking tasks for synthetic biology gene regulatory networks.  

\subsection{Related Works}

In model-free reinforcement learning, stability is rarely addressed due to the formidable challenge of analyzing and designing the closed-loop system dynamics in a model-free manner~\cite{bucsoniu2018reinforcement}, and the associated stability theory in model-free RL remains as an open problem~\cite{bucsoniu2018reinforcement,gorges2017relations}.

Recently, Lyapunov analysis is used in model-free RL to solve control problems with safety constraints \cite{chow2018lyapunov,chow2019lyapunov}. In~\cite{chow2018lyapunov}, the Lyapunov-based approach for solving constrained Markov decision processes is proposed with a novel way of constructing the Lyapunov function through linear programming. In~\cite{chow2019lyapunov}, the above results were further generalized to continuous control tasks. Even though Lyapunov-based methods were adopted in these results, neither of them addressed the stability of the system. In this paper, the sufficient conditions for a dynamic system being stable are derived. Furthermore, it is shown that these conditions can be verified through sampling and ensured through model-free learning.

Other interesting results on the stability of learning-based control systems are reported in recent years. In~\cite{postoyan2017stability}, an initial result is proposed for the stability analysis of deterministic nonlinear systems with optimal controller for infinite-horizon discounted cost, based on the assumption that discount is sufficiently close to $1$. However, in practice, it is rather difficult to guarantee the optimality of the learned policy unless certain assumptions on the system dynamics are made \cite{murray2003adaptive}. Furthermore, the exploitation of multi-layer neural networks as function approximators \cite{lillicrap2015continuous} only adds to the impracticality of this requirement. \textcolor{black}{In this paper, it is shown that it is sufficient to ensure stability by satisfying the Lyapunov criterion that is evaluated on samples and thus one is exempt from finding the optimal/suboptimal solutions.} In~\cite{berkenkamp2017safe}, local stability of Lipschitz continuous dynamic systems is analyzed by validating the ``energy decreasing'' condition on discretized points in the subset of state space with the help of a learned model (Gaussian process). Nevertheless, the discretization technique may become infeasible as the dimension and space of interest increases, limiting its application to rather simple and low-dimensional systems. \textcolor{black}{In this paper, the proposed method is applicable to the general class of stochastic dynamic systems modeled by MDP and does not need to learn a model for stability analysis and controller design.}

\section{PROBLEM STATEMENT}
In this paper, we focus on the stabilization and tracking tasks for systems modeled by Markov decision process (MDP). The state of a robot and its environment at time $t$ is given by the state $s_t\in\mathcal{S}\subseteq \mathbb{R}^n$, where $\mathcal{S}$ denotes the state space. The robot then takes an action $a_t \in\mathcal{A}\subseteq \mathbb{R}^m$ according to a stochastic policy $\pi(a_t|s_t)$, resulting in the next state $s_{t+1}$. The transition of the state is modeled by the transition probability $P (s_{t+1}|s_t,a_t)$. In both stabilization and tracking tasks, there always is a cost function $c(s_t,a_t)$ to measure how good or bad a state-action pair is.

In stabilization tasks, the goal is to find a policy $\pi$ such that the norm of state $\Vert s_{t} \Vert$ goes to zero eventually, where $\Vert \cdot \Vert$ denotes the Euclidean norm. In this case, cost function $c(s_t,a_t) = \mathbb{E}_{P(\cdot|s_t,a_t)}\Vert s_{t+1} \Vert$. In tracking tasks, we divide the state $s$ into two vectors, $s^1$ and $s^2$, where $s^1$ is composed of elements of $s$ that are aimed at tracking the reference signal $r$, while $s^2$ contains the rest. The reference signal could be the desired velocity, path and even the picture of grasping an object in a certain pose. For tracking tasks, $c(s_t,a_t) = \mathbb{E}_{P(\cdot|s_t,a_t)}\Vert s_{t+1}^1-r \Vert$.

From a control perspective, both stabilization and tracking tasks are related to the asymptotic stability of the closed-loop system (or error system) under $\pi$, i.e., starting from an initial point, the trajectories of state always converge to the origin or reference trajectory. Let $c_\pi(s_t)\define\mathbb{E}_{a\sim\pi}c(s_t,a_t)$ denote the cost function under the policy $\pi$, the definition of stability studied in this paper is given as follows.
\begin{definition}
\label{def:mss}
The stochastic system is said to be stable in mean cost if $\lim_{t\rightarrow \infty }\mathbb{E}_{s_{t}} c_\pi(s_{t})=0$ holds for any initial condition $s_{0}\in \{s_{0}|c_\pi(s_{0})\leq b\}$. If $b$ is arbitrarily large then the stochastic system is globally stable in mean cost.
\end{definition}
The above definition is equivalent to the mean square stability \cite{shaikhet1997necessary,bolzern2010markov} when the cost $c$ is chosen to be the norm of the state; it is also equivalent to the partial stability \cite{vorotnikov2005partial,haddad2015finite} when $c(s_t,a_t) = \mathbb{E}_{P(\cdot|s_t,a_t)}\Vert s_{t+1}^1-r \Vert$. Thus the stabilization and tracking tasks can be collectively summarized as finding a policy $\pi$ such that the closed-loop system is stable in mean cost according to Definition~\ref{def:mss}.

Before proceeding, some notations are to be defined. $\rho (s_0)$ denotes the distribution of starting states. The closed-loop transition probability is denoted as $P_\pi(s'|s)\define\int_\mathcal{A}\pi(a|s)P(s'|s,a)\mathrm{d}a$. We also introduce the closed-loop state distribution at a certain instant $t$ as $P(s|\rho, \pi, t)$, which could be defined iteratively: $P(s'|\rho, \pi, t+1) =\int_\mathcal{S} P_\pi(s'|s)P(s|\rho, \pi, t)\mathrm{d}s, \forall t\in\mathbb{Z}_{+}$ and $P(s|\rho, \pi, 0) =\rho(s)$.

\section{DATA-BASED STABILITY ANALYSIS}\label{sec:main results}
In this section, we propose the main assumptions and a new theorem for stability analysis of stochastic systems.
We assume that the Markov chain induced by policy $\pi$ is ergodic with a unique stationary distribution $q_\pi$, 
\vspace{-0.05cm}
$$q_\pi(s)=\lim_{t\rightarrow\infty}P(s|\rho,\pi,t)$$ 
as commonly exploited by many RL literature \cite{sutton2009convergent, korda2015td, bhandari2018finite, zou2019finite}.

In Definition~\ref{def:mss}, stability is defined in relation to the set of starting states, which is also called the region of attraction (ROA). If the MSS system starts within the ROA, its trajectory will be surely attracted to the equilibrium. To build a data-based stability guarantee, we need to ensure that the states in ROA are accessible for the stability analysis. Thus the following assumption is made to ensure that every state in ROA has a chance to be sampled as the starting state. 

\begin{assumption}\label{initial state assumption}
There exists a positive constant $b$ such that $\rho(s)> 0, \forall s\in\{s|c_\pi(s)\leq b\}$.
\end{assumption}

Our approach is to construct/find a Lyapunov function $L: \mathcal{S} \to \mathbb{R}_+$ of which the difference along the state trajectory is negative definite, so that the state goes in the direction of decreasing the value of Lyapunov function and eventually converges to the origin. The Lyapunov's method has long been used for stability analysis and controller design in control theory~\cite{boukas2000robust}, but mostly exploited along with a \emph{known} model so that the energy decreasing condition on the entire state space could be transformed into one inequality regarding model parameters~\cite{slotine1991applied, sastry2013nonlinear}. In the following, we show that without a dynamic model, this ``infinity'' problem could be solved through sampling and sufficient conditions for a stochastic system to be stable in mean cost are given.
\begin{theorem}\label{them:MSS}
The stochastic system is stable in mean cost if there exists a function $L:\mathcal{S}\rightarrow \mathbb{R}_{+}$\ and positive constants $\alpha _{1}$, $\alpha _{2}$ and $\alpha_{3}$, such that%
\begin{align}
\begin{split}\label{Theorem 2-1}
\ \ \ \ \ \ \ \ \ \alpha_{1}c_\pi\left( s\right) \leq L(s)\leq \alpha _{2}c_\pi\left( s\right)
\end{split}\\
\begin{split}\label{Theorem 2-2}
\mathbb{E}_{s\sim \mu_\pi }(\mathbb{E}_{s^{\prime }\sim P_{\pi }}L(s^{\prime
})-L(s))\leq -\alpha_{3} \mathbb{E}_{s\sim \mu_\pi }c_\pi\left( s\right)
\end{split}
\end{align}
where 
\vspace{-0.3cm}
$$\mu_\pi(s)\define \lim_{N\rightarrow\infty}\frac{1}{N}\sum_{t=0}^N P(s_t=s|\rho,\pi,t)$$ is the (infinite) sampling distribution. 
\end{theorem}

\begin{proof}
The existence of the sampling distribution $\mu_\pi(s)$ is guaranteed by the existence of $q_\pi(s)$. Since the sequence $\{P(s|\rho,\pi,t), t\in\mathbb{Z}_+\}$ converges to $q_\pi(s)$ as $t$ approaches $\infty$, then by the Abelian theorem, the sequence $\{\frac{1}{N}\sum_{t=0}^N P(s|\rho,\pi,t), N\in\mathbb{Z}_+\}$ also converges and $\mu_\pi(s) = q_\pi(s)$. Combined with the form of $\mu_\pi$, (\ref{Theorem 2-2}) infers that
\vspace{-0.1cm}
\begin{equation}
\begin{aligned}
    &\int_\mathcal{S}\lim_{N\rightarrow\infty}\frac{1}{N}\sum_{t=0}^N P(s|\rho,\pi,t)(\mathbb{E}_ {P_{\pi}(s^{\prime}|s)}L(s^{\prime})-L(s))\mathrm{d}s \\
    \leq& -\alpha_{3} \mathbb{E}_{s\sim q_\pi}c_\pi\left(s\right)
\end{aligned}
\label{proof:theorem MSS 1}
\end{equation}
First, on the left-hand-side, $L(s)\leq \alpha_2 c_\pi(s)$ for all $s\in\mathcal{S}$ according to (\ref{Theorem 2-1}). Since the probability density function $P(s|\rho,\pi,t)$ is (assumed to be) a bounded function on $\mathcal{S}$ for all $t$, thus there exists a constant $M$ such that
\begin{equation*}
 P(s|\rho,\pi,t)L(s)\leq M\alpha_2 c_\pi\left(s\right), \forall s\in \mathcal{S}, \forall t\in \mathbb{Z}_+
\end{equation*}
Second, the sequence $\{\frac{1}{N}\sum_{t=0}^N P(s|\rho,\pi,t)L(s), N\in\mathbb{Z}_+\}$ converges point-wise to the function $q_\pi(s)L(s)$. According to the Lebesgue's Dominated convergence theorem \cite{royden1968real}, if a sequence ${f_n(s)}$  converges point-wise to a function $f$ and is dominated by some integrable function $g$ in the sense that,
\begin{equation*}
\vert f_n(s) \vert \leq g(s), \forall s\in \mathcal{S},\forall n
\end{equation*}
Then we get
\vspace{-0.2cm}
\begin{equation*}
 \lim_{n\rightarrow\infty}\int_\mathcal{S}f_n(s)\mathrm{d}s= \int_\mathcal{S}\lim_{n\rightarrow\infty}f_n(s)\mathrm{d}s
\end{equation*}
Thus the left-hand-side of (\ref{proof:theorem MSS 1})
\begin{equation*}
\begin{aligned}
 &\int_\mathcal{S}\lim_{N\rightarrow\infty}\frac{1}{N}\sum_{t=0}^N P(s|\rho,\pi,t)(\int_\mathcal{S} P_{\pi}(s^{\prime}|s)L(s^{\prime})\mathrm{d}s^{\prime}-L(s))\mathrm{d}s\\
 =&\lim_{N\rightarrow\infty}\frac{1}{N}(\sum_{t=1}^{N+1} \mathbb{E}_{P(s|\rho,\pi,t)}L(s)-\sum_{t=0}^{N} \mathbb{E}_{P(s|\rho,\pi,t)}L(s))\\
 =&\lim_{N\rightarrow\infty}\frac{1}{N} \left( \mathbb{E}_{ P(s|\rho,\pi,N+1)}L(s)-\mathbb{E}_{\rho(s)}L(s)\right)
\end{aligned}
\end{equation*}

Thus taking the relations above into consideration, (\ref{proof:theorem MSS 1}) infers
\vspace{-0.2cm}
\begin{equation}
\begin{aligned}
&\lim_{N\rightarrow\infty}\frac{1}{N} \left( \mathbb{E}_{P(s|\rho,\pi,N+1)}L(s)-\mathbb{E}_{ \rho(s)}L(s)\right) \\
\leq &-\alpha_3\lim_{t\rightarrow\infty}\mathbb{E}_{P(s|\rho,\pi,t)}c_\pi\left(s\right)
\end{aligned}
\end{equation}
Since $\mathbb{E}_{\rho(s)}L(s)$ is a finite value and $L$ is positive definite, it follows that
\vspace{-0.1cm}
\begin{equation}
\lim_{t\rightarrow\infty}\mathbb{E}_{ P(s|\rho,\pi,t)}c_\pi\left(s\right)\leq \lim_{N\rightarrow\infty}\frac{1}{N} (\frac{1}{\alpha_3}\mathbb{E}_{\rho(s)}L(s))
= 0 \label{final eq}
\end{equation}
Suppose that there exists a state $s_0\in\{s_0|c_\pi(s_0)\leq b\}$ and a positive constant $d$ such that
$\lim_{t\rightarrow\infty}\mathbb{E}_{ P(s|s_0,\pi,t)}c_\pi\left(s\right)=d$, or $\lim_{t\rightarrow\infty}\mathbb{E}_{ P(s|s_0,\pi,t)}c_\pi\left(s\right)=\infty$. Since $\rho(s_0)> 0$ for all starting states in $\{s_0|c_\pi(s_0)\leq b\}$ (Assumption~\ref{initial state assumption}), it follows that $\lim_{t\rightarrow\infty}\mathbb{E}_{s_t\sim P(\cdot|\pi,\rho)}c_\pi\left(s_t\right)>0$, which is contradictory with (\ref{final eq}). Thus $\forall s_0\in \{s_0|c_\pi(s_0)\leq b\}$, $\lim_{t\rightarrow\infty}\mathbb{E}_{ P(s|s_0,\pi,t)}c_\pi\left(s\right)=0$.
Thus the system is stable in mean cost by Definition~\ref{def:mss}.
\end{proof}

(\ref{Theorem 2-1}) directs the choice and construction of Lyapunov function, of which the details are deferred to Section~\ref{sec:algorithm}. (\ref{Theorem 2-2}) is called the energy decreasing condition and is the major criterion for determining stability. 

\begin{remark}
This remark is on the connection to previous results concerning the stability of stochastic systems.
It should be noted that the stability conditions of Markov chains have been reported in~\cite{shaikhet1997necessary, meyn2012markov}, however, of which the validation requires verifying infinite inequalities on the state space if $\mathcal{S}$ is continuous. On the contrary, our approach solely validates one inequality (\ref{Theorem 2-2}) related to the sampling distribution $\mu$, which further enables data-based stability analysis and policy learning. 
\end{remark}

\section{ALGORITHM}
\label{sec:algorithm}

In this section, we propose an actor-critic RL algorithm to learn stability guaranteed policies for the stochastic system. First, we introduce the Lyapunov critic function $L_c$ and show how it is constructed. Then based on the maximum entropy actor-critic framework, we use the Lyapunov critic function in the policy gradient formulation.

\subsection{Lyapunov Critic Function}

In our framework, the Lyapunov critic $L_c$ plays a role in both stability analysis and the learning of the actor. To enable the actor-critic learning, the Lyapunov critic is designed to be dependent on $s$ and $a$ and satisfies $L(s) = \mathbb{E}_{a\sim \pi} L_c(s,a)$ with the Lyapunov function $L(s)$, such that it can be exploited in judging the value of \eqref{Theorem 2-2}. In view of the requirement above, $L_c$ should be a non-negative function of the state and action, $L_c:\mathcal{S}\times\mathcal{A}\rightarrow \mathbb{R}_+$. In this paper, we construct Lyapunov critic with the following parameterization technique,
\begin{equation}\label{eq:lyapunov critic parameterization}
    L_c(s,a)=f_\phi(s,a)^T f_\phi(s,a)
\end{equation}
where $f_\phi$ is the output vector of a fully connected neural network with parameter $\phi$. \textcolor{black}{This parameterization ensures the positive definiteness of $L_c(s,a)$, which is necessary since $L(s)$ is positive definite according to \eqref{Theorem 2-1} and $L(s)$ is the expectation of $L_c(s,a)$ over the distribution of actions.}

Theoretically, some functions, such as the norm of state and value function, naturally satisfy the basic requirement of being a Lyapunov function \eqref{Theorem 2-1}. These functions are referred to as Lyapunov candidates. However, Lyapunov candidates are conceptual functions without any parameterization, thus their gradient with respect to the controller is intractable and are not directly applicable in an actor-critic learning process. In the proposed framework, the Lyapunov candidate acts as a supervision signal during the training of $L_c$. During training, $L_c$ is updated to minimize the following objective function,
\vspace{-0.2cm}
\begin{equation}\label{eq:critic objective}
    J(L_c) = \mathbb{E}_{ \mathcal{D}}\left[\frac{1}{2}(L_c(s,a)-L_{\text{target}}(s,a))^2\right]
\end{equation}
where $L_{\text{target}}$ is the approximation target related to the chosen Lyapunov candidate, \textcolor{black}{$L(s) = \mathbb{E}_{a\sim \pi} L_{\text{target}}(s,a)$} and $\mathcal{D}$ is the set of collected transition pairs. In~\cite{chow2018lyapunov} and~\cite{berkenkamp2017safe}, the value function has been proved to be a valid Lyapunov candidate where the approximation target is 
\vspace{-0.1cm}
\begin{equation}
L_{\text{target}}(s,a) = c + \max_{a'}\gamma L'_c(s', a')
\vspace{-0.1cm}
\end{equation}
where $L'_c$ is the target network parameterized by $\phi^{\prime}$ as typically used in the actor-critic methods~\cite{haarnoja2018soft,lillicrap2015continuous}. $L'_c$ has the same structure as $L_c$, but the parameter $\phi^{\prime}$ is updated through exponentially moving average of weights of $L_c$ controlled by a hyperparameter $\tau\in \mathbb{R}_{(0,1)}$, $\phi^{\prime}_{k+1}\leftarrow \tau\phi_{k} + (1-\tau)\phi^{\prime}_{k}$.

In addition to value function, the sum of cost over a finite time horizon could also be employed as Lyapunov candidate, which is exploited in model predictive control literature~\cite{mayne1990receding,mayne2000constrained} for stability analysis. In this case,
\vspace{-0.2cm}
\begin{equation}
L_{\text{target}}(s,a) = \sum_t^{t+N}\mathbb{E}c_t
\vspace{-0.1cm}
\end{equation}
Here, the time horizon $N$ is a hyperparameter to be tuned, of which the influence will be demonstrated in the experiment in Section~\ref{sec:experiment}.

The choice of the Lyapunov candidate plays an important role in learning a policy. Value function evaluates the infinite time horizon and thus offers a better performance in general but is rather difficult to approximate because of significant variance and bias~\cite{schulman2015high}. On the other hand, the finite horizon sum of cost provides an explicit target for learning a Lyapunov function, thus inherently reduces the bias and enhances the learning process. However, as the model is unknown, predicting the future costs based on the current state and action inevitably introduces variance, which grows as the prediction horizon extends. In principle, for tasks with simple dynamics, the sum-of-cost choice enhances the convergence of learning and robustness of the trained policies, while for complicated systems the choice of value function generally produces better performance. In this paper, we use both value function and sum-of-cost as Lyapunov candidates. Later in Section~\ref{sec:experiment}, we will show the influence of these different choices upon the performance and robustness of trained policies.

\subsection{Lyapunov Actor-Critic Algorithm}

In this subsection, we will focus on how to learn the controller in a novel actor-critic framework called Lyapunov Actor-Critic (LAC), such that the inequality \eqref{Theorem 2-2} is satisfied. The policy learning problem is summarized as the following constrained optimization problem,
\vspace{-0.1cm}
\begin{align}
    \text{find }&\pi_\theta\\
    \text{s.t. }& \mathbb{E}_{\mathcal{D}}(L_c(s',f_\theta(\epsilon,s'))-L_c(s,a)+\alpha_3c) \label{eq:parameterized lyapunov inequality} \\
    &-\mathbb{E}_{\mathcal{D}}\log(\pi(a|s))\geq\mathcal{H}_t \label{eq:maximum entropy}
\end{align}
\hspace{-0.2cm}
where the stochastic policy $\pi_\theta$ is parameterized by a deep neural network $f_\theta$ that is dependent on $s$ and a Gaussian noise $\epsilon$. The constraint \eqref{eq:parameterized lyapunov inequality} is the parameterized inequality \eqref{Theorem 2-2}, which can be evaluated through sampling. One may be curious why in the second term of \eqref{LAC}, only one Lyapunov critic is explicitly dependent on the stochastic policy, while the other dependent on the samples of the action. First, note that this estimator is also unbiased estimation of \eqref{Theorem 2-2}, although the variance may be increased compared to replacing $a$ with $f_\theta(s)$. From a more practical perspective, having the second Lyapunov critic explicitly dependent on $\theta$ will introduce a term in the policy gradient that updates $\theta$ to increase the value of $L(s)$, which is contradictory to our goal of stabilization. \eqref{eq:maximum entropy} is the minimum entropy constraint borrowed from the maximum entropy RL framework to improve the exploration in the action space \cite{haarnoja2018soft}, and $\mathcal{H}_t$ is the desired bound. Solving the above constrained optimization problem is equivalent to minimizing the following objective function,
\begin{align}
\begin{aligned}
 J(\theta) =& \mathbb{E}_{(s,a,s',c)\sim\mathcal{D}}[ \beta (\log(\pi_\theta(f_\theta(\epsilon,s)|s))+\mathcal{H}_t)\\ &+\lambda(L_c(s',f_\theta(\epsilon,s'))-L_c(s,a)+
    \alpha_3c)] 
\end{aligned}
\label{LAC}
\end{align}
where $\beta$ and $\lambda$ are Lagrange multipliers that control the relative importance of constraint\eqref{eq:parameterized lyapunov inequality} and \eqref{eq:maximum entropy}.

In the actor-critic framework, the parameters of the policy network are updated through stochastic gradient descent of \eqref{LAC}, which is approximated by 
\begin{equation}
\begin{aligned}
\nabla_\theta J(\theta)
= &\beta \nabla_\theta  \log(\pi_\theta(a|s))+ \\ 
& \beta \nabla_a  \log(\pi_\theta(a|s))\nabla_\theta f_\theta(\epsilon,s) +\\
& \lambda\nabla_{a'}L_c(s',a')\nabla_\theta f_\theta(\epsilon,s')
\end{aligned}
\label{LAC_policy_gradient}
\end{equation}

The value of Lagrange multipliers $\beta$ and $\lambda$ are adjusted by gradient ascent to maximize the following objectives respectively while being clipped to be positive, 
\begin{gather*}
J(\beta) = \beta\mathbb{E}_{\mathcal{D}}  [\log\pi_\theta( a|s)+\mathcal{H}_t]\label{eq:temperature update}\\
J(\lambda) =\lambda\mathbb{E}_{\mathcal{D}}[ L_c(s^{\prime },f_\theta(s^{\prime},\epsilon)) -L_c(s,a) +\alpha_3c]
\end{gather*}
During training, the Lagrange multipliers are updated by 
\begin{align*}
    \lambda &\leftarrow \max (0,\lambda +  \delta \nabla_\lambda J(\lambda)) \notag \\
    \beta &\leftarrow \max (0,\beta +  \delta \nabla_{\beta} J(\beta)) \notag 
\end{align*}
where $\delta$ is the learning rate.
The pseudo-code of the proposed algorithm is shown in Algorithm~\ref{algo:LAC}.

\begin{algorithm}[H]
   \caption{Lyapunov-based Actor-Critic (LAC)}
   \label{algo:LAC}
\begin{algorithmic}
    \REQUIRE
    Maximum episode length $N$ and maximum update steps $M$
   \REPEAT
   \STATE Sample $s_0$ according to $\rho$
   \FOR{$t=1$ to $N$}
   \STATE Sample $a$ from $\pi_\theta(a|s)$ and step forward
   \STATE Observe $s^\prime$, $c$ and store $(s,a,c,s^\prime)$ in $\mathcal{D}$
   \ENDFOR
   \FOR{$i=1$ to $M$}
    \STATE Sample mini-batches of transitions from $\mathcal{D}$ and update $L_c$, $\pi$, Lagrange multipliers with \eqref{eq:critic objective}, \eqref{LAC_policy_gradient}
   \ENDFOR
   \UNTIL{\eqref{eq:parameterized lyapunov inequality} is satisfied}

\end{algorithmic}

\end{algorithm}

\begin{figure}[htb]
    \centering
    \subfigure[CartPole]{
    \includegraphics[width=0.45\columnwidth]{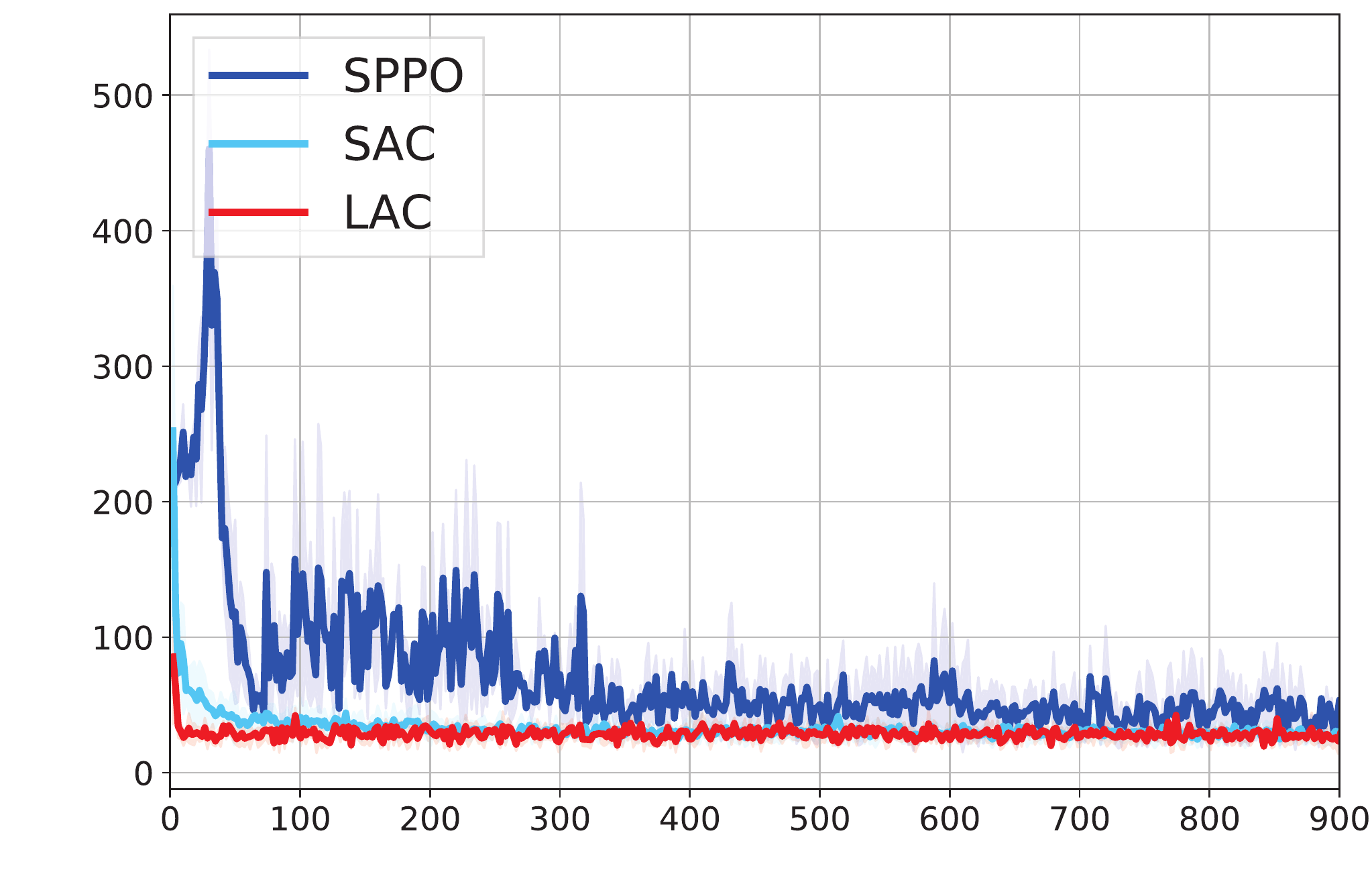}
    }
    \centering
    \subfigure[HalfCheetah]{
    \includegraphics[width=0.45\columnwidth]{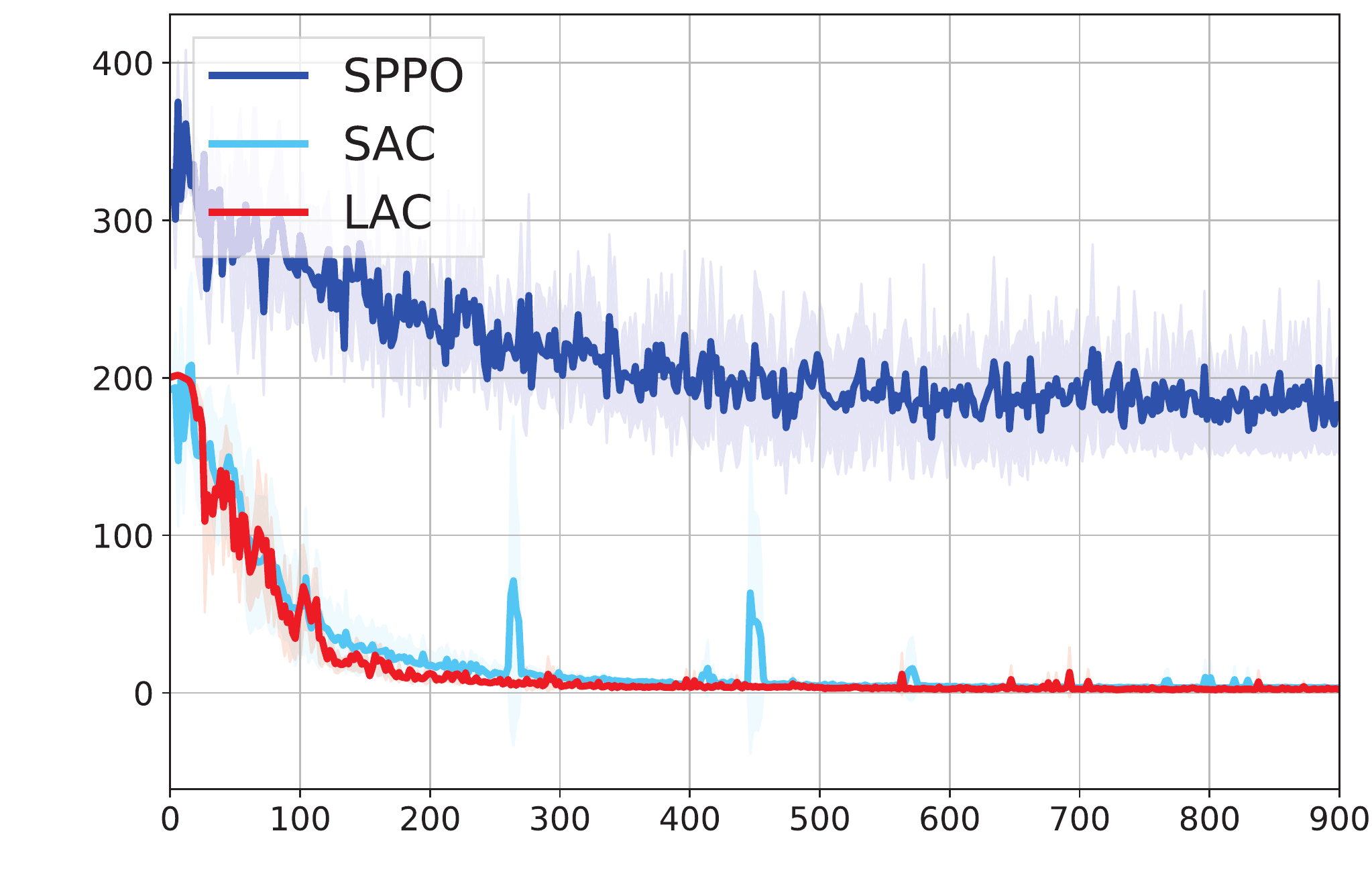}
    }
    \centering
    \subfigure[FetchReach]{
    \includegraphics[width=0.45\columnwidth]{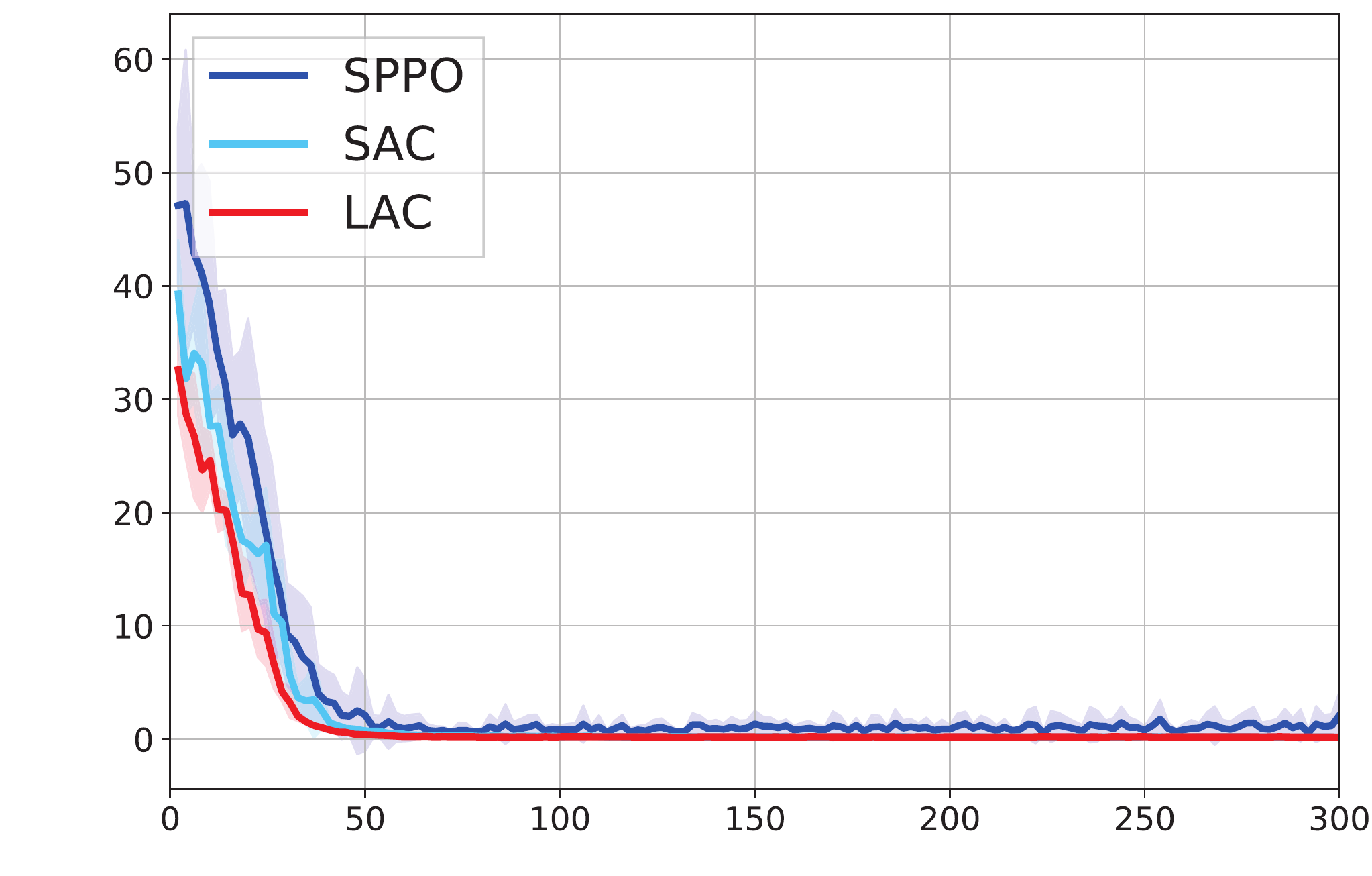}
    }    
    \centering
    \subfigure[GRN]{
    \includegraphics[width=0.45\columnwidth]{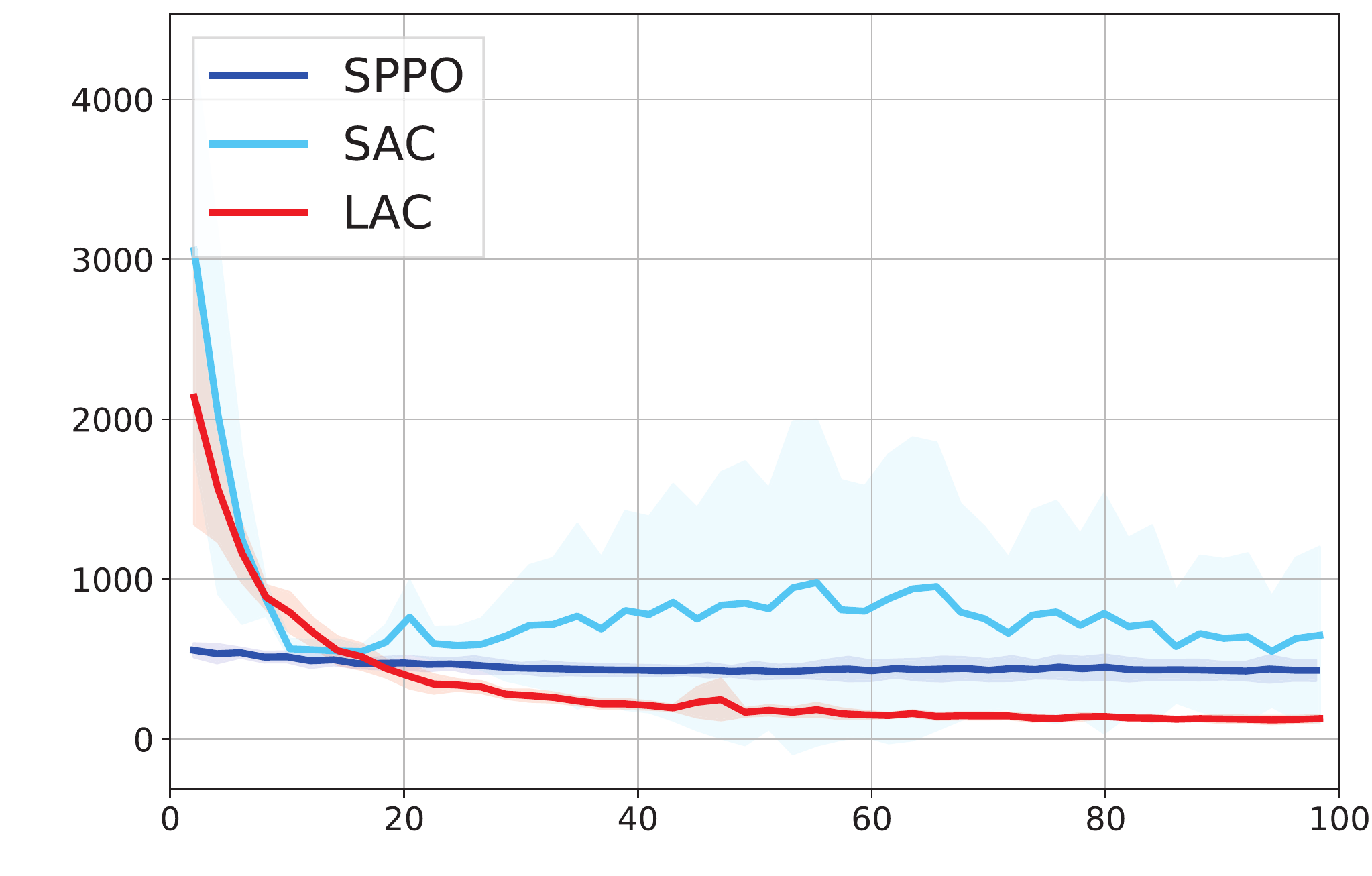}
    }
    \centering
    \subfigure[Swimmer]{
    \includegraphics[width=0.45\columnwidth]{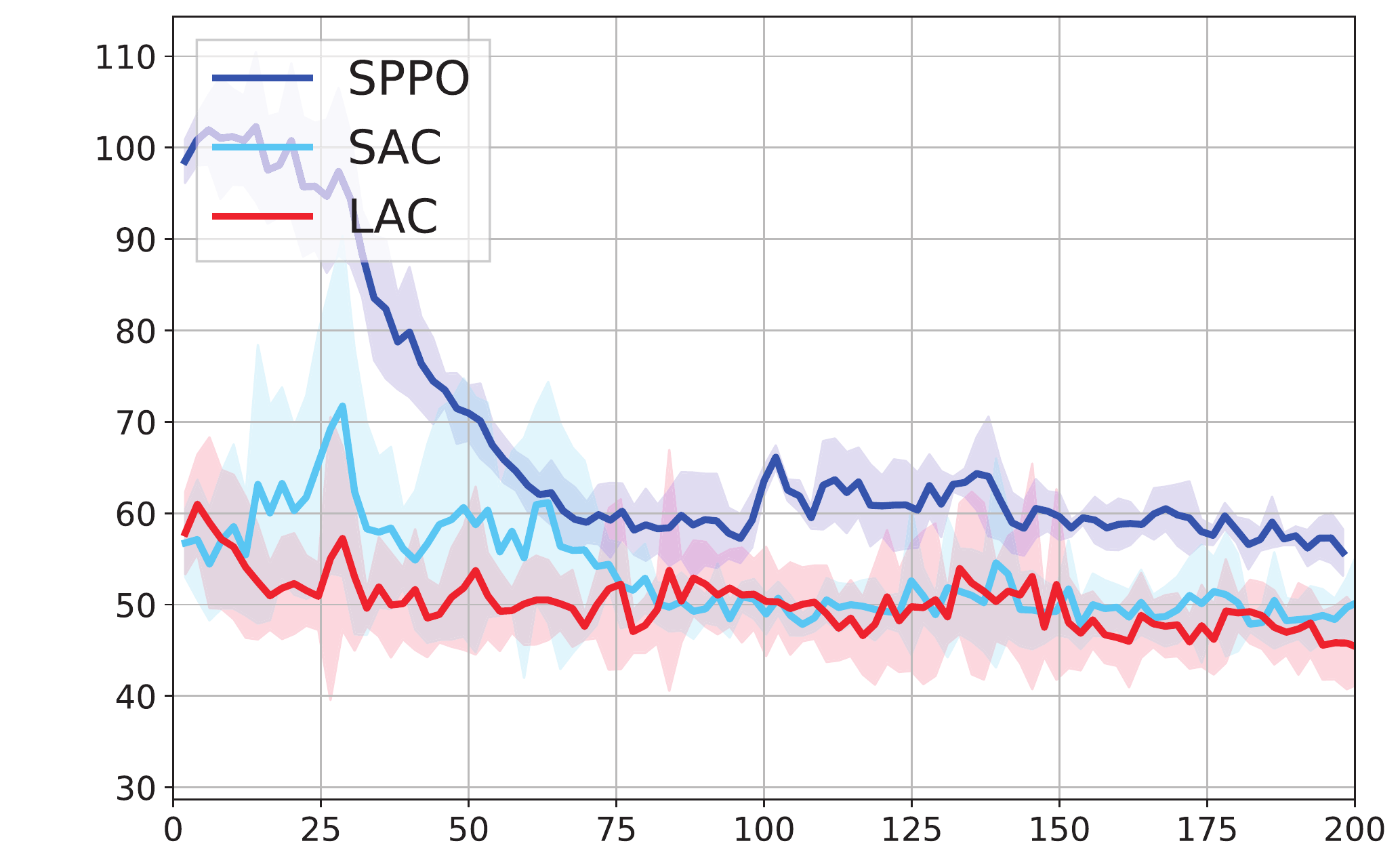}
    \label{Swimmer}
    }
    \centering
    \subfigure[Minitaur]{
    \includegraphics[width=0.46\columnwidth]{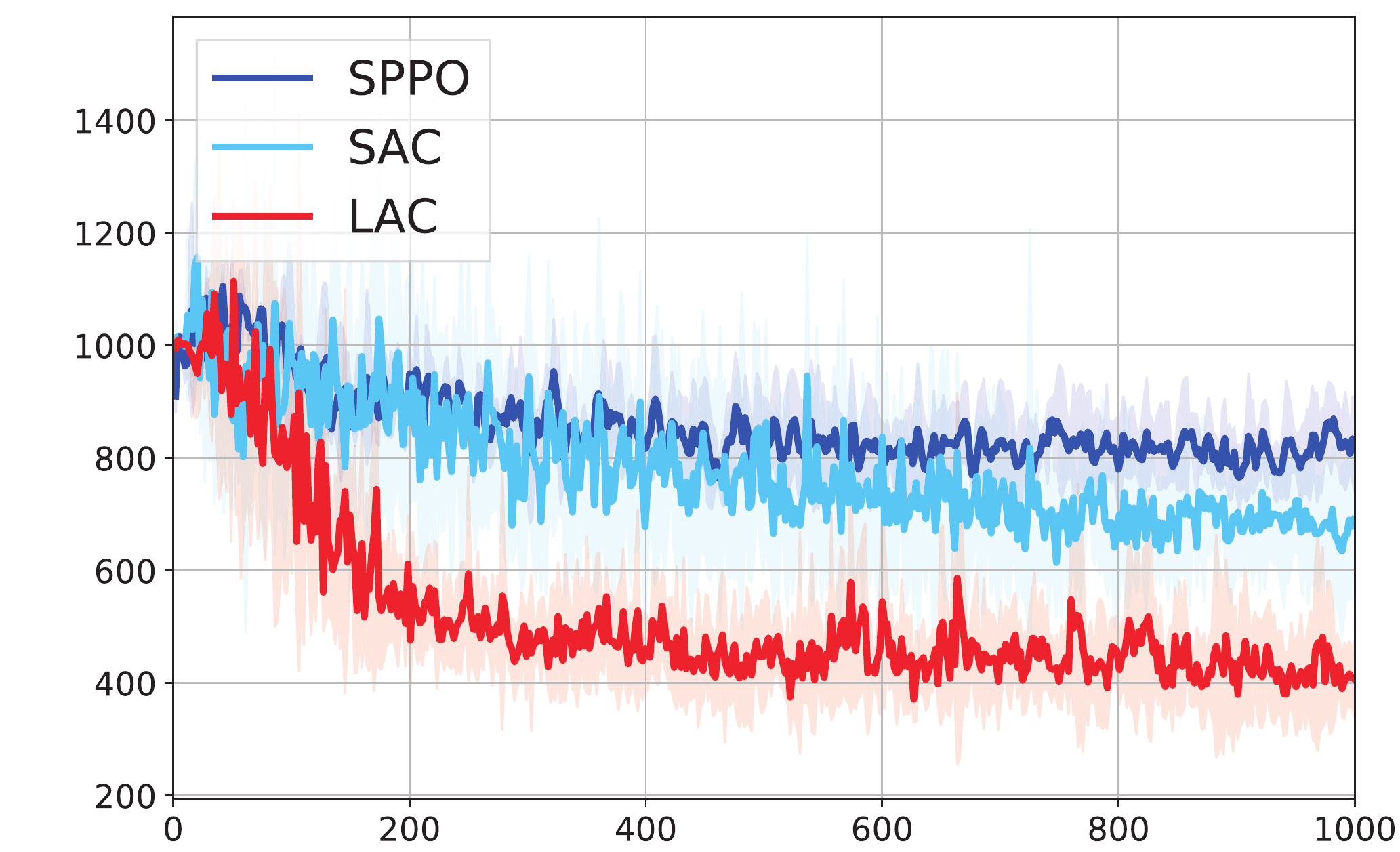}
    \label{Minitaur}
    }
    \caption{Cumulative control performance comparison. 
    The Y-axis indicates the total cost during one episode and the X-axis indicates the total time steps in thousand. The shadowed region shows the 1-SD confidence interval over 10 random seeds. Across all trials of training, LAC converges to stabilizing solution with comparable or superior performance compared with SAC and SPPO.}
    \label{fig:performance}
\end{figure}

\section{EXPERIMENT}\label{sec:experiment}
In this section, we illustrate five simulated robotic control problems to demonstrate the general applicability of the proposed method.
First of all, the classic control problem of CartPole balancing from control and RL literature~\cite{barto1983neuronlike} is illustrated. Then, we consider more complicated high-dimensional continuous control problems of 3D robots, e.g., HalfCheetah and FetchReach, using MuJoCo physics engine~\cite{todorov2012mujoco}, a multi-joint Swimmer robot \cite{tassa2018deepmind}, and a full quadruped (Minitaur) simulated by PyBullet platform \cite{coumans2016pybullet}. Last, we extend our approach to control autonomous systems in the cell, i.e., molecular synthetic biological gene regulatory networks (GRN). Specifically, we consider the problem of reference tracking for two GRNs~\cite{elowitz2000synthetic}.

The proposed method is evaluated for the following aspects:

\begin{itemize}
    \item Convergence: does the proposed training algorithm converge with random parameter initialization and does the stability condition \eqref{Theorem 2-2} hold for the learned policies;
    \item Performance: can the goal of the task be achieved or the cumulative cost be minimized;
    \item Stability: if \eqref{Theorem 2-2} hold, are the closed-loop systems stable indeed and generating stable state trajectories;
    \item Robustness: how do the trained policies perform when faced with uncertainties unseen during training, such as parametric variation and external disturbances;
    \item Generalization: can the trained policies generalize to follow reference signals that are different from the one seen during training.
\end{itemize}

We compare our approach with soft actor-critic (SAC)~\cite{haarnoja2018soft}, one of the state-of-the-art actor-critic algorithms that outperform a series of RL methods such as DDPG~\cite{ddpg}, PPO~\cite{schulman2017proximal} on the continuous control benchmarks. 
The variant of safe proximal policy optimization (SPPO)~\cite{chow2019lyapunov}, a Lyapunov-based method, is also included in the comparison. The original SPPO is developed to deal with constrained MDP, where safety constraints exist. In our experiments, we modify it to apply the Lyapunov constraints on the MDP tasks and see whether it can achieve the same stability guarantee as LAC. In CartPole, we also compare with the linear quadratic regulator (LQR), a classical model-based optimal control method for stabilization. For both algorithms, the hyperparameters are tuned to reach their best performance.

The outline of this section is as follows. In Section~\ref{sec:exp:training}, the performance, and convergence of LAC are demonstrated and compared with SAC. In Section~\ref{sec:exp:stability evaluation}, a straight forward demonstration of stability is made by comparing with the baseline method. In Section~\ref{sec:exp:evaluation}, the ability of generalization and robustness of the trained policies are evaluated and analyzed. Finally, in Section~\ref{sec:exp:candidates}, we show the influence of choosing different Lyapunov candidates upon the performance and robustness of trained policies. 

The hyperparameters of LAC and the detailed experiment setup are deferred to Appendix \cite{han2020actor}. The code for reproduction can be found in our GitHub repository \footnote{\url{https://github.com/hithmh/Actor-critic-with-stability-guarantee}}.

\subsection{Performance}\label{sec:exp:training}

In each task, both LAC, SAC, and SPPO are trained 10 times with random initialization, average total cost, and its variance during training are demonstrated in \autoref{fig:performance}.
In the examples (a)-(c) and (e), SAC and LAC perform comparably in terms of the total cost at convergence and the speed of convergence, while SPPO could converge in Cartpole, FetcheReach, and Swimmer. In GRN and CompGRN (see \autoref{fig:performance}(d) and Fig. S9(b) in the supplementary material), SAC is not always able to find a policy that is capable of completing the control objective, resulting in the bad average performance. In the Minitaur example (see \autoref{fig:performance}(f)), SAC and SPPO can only converge to suboptimal solutions. On the contrary, LAC performs stably regardless of the random initialization. 
As shown in \autoref{fig:performance}, LAC converges stably in all experiments.

\subsection{Evaluation of Stability}\label{sec:exp:stability evaluation}
In this part, further comparison between the stability-assured method (LAC) and that without such guarantee (SAC) is made, by demonstrating the closed-loop system dynamic with the trained policies. A distinguishing feature of the stability assured policies is that it can force and sustain the state or tracking error to zero. This could be intuitively demonstrated by the state trajectories of the closed-loop system.

\begin{figure}[htb]
    \vspace{-0.15cm}
    \centering
    \subfigure[LAC-GRN]{
    \includegraphics[width=0.42\columnwidth]{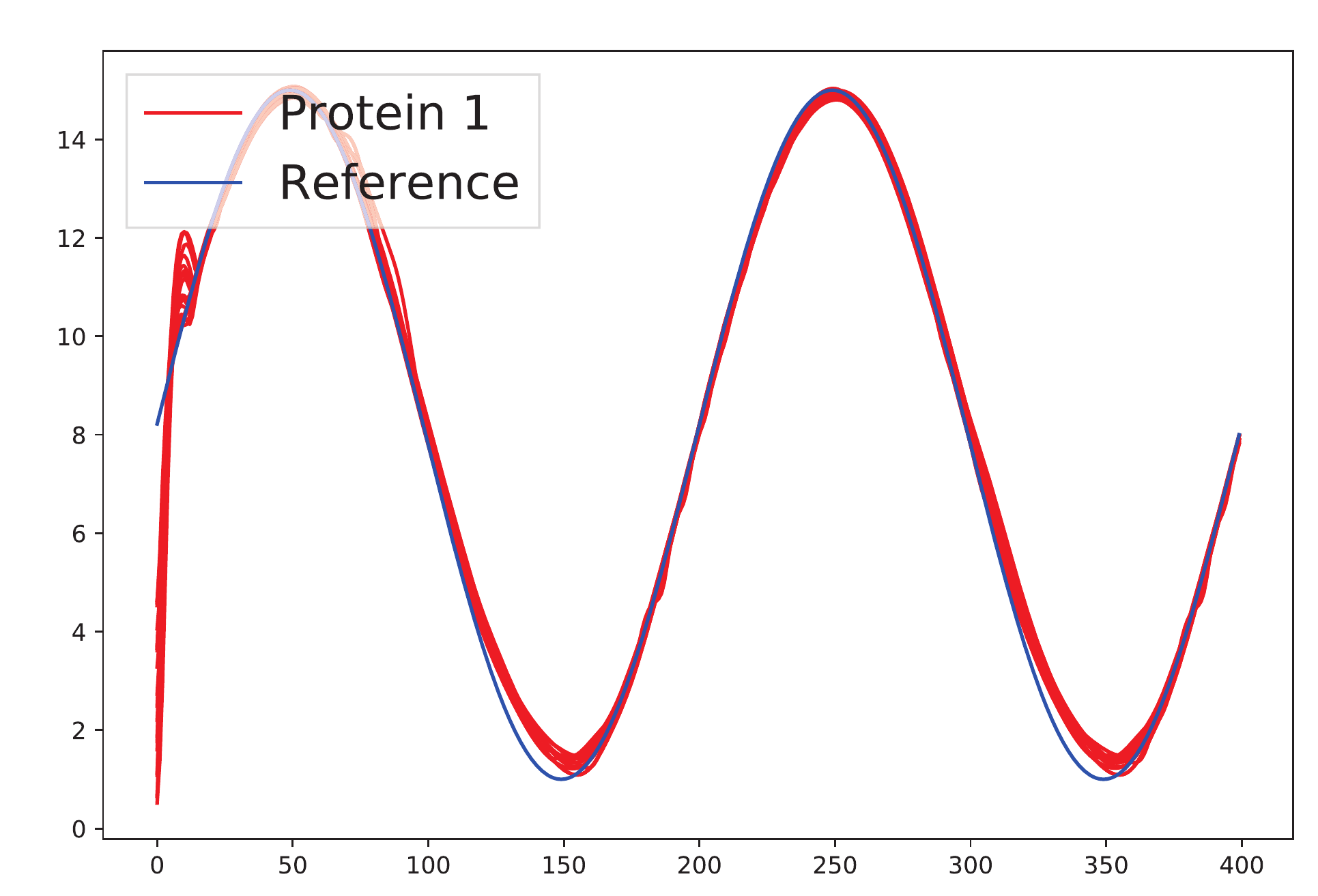}
    }
    \centering
    \subfigure[SAC-GRN]{
    \includegraphics[width=0.42\columnwidth]{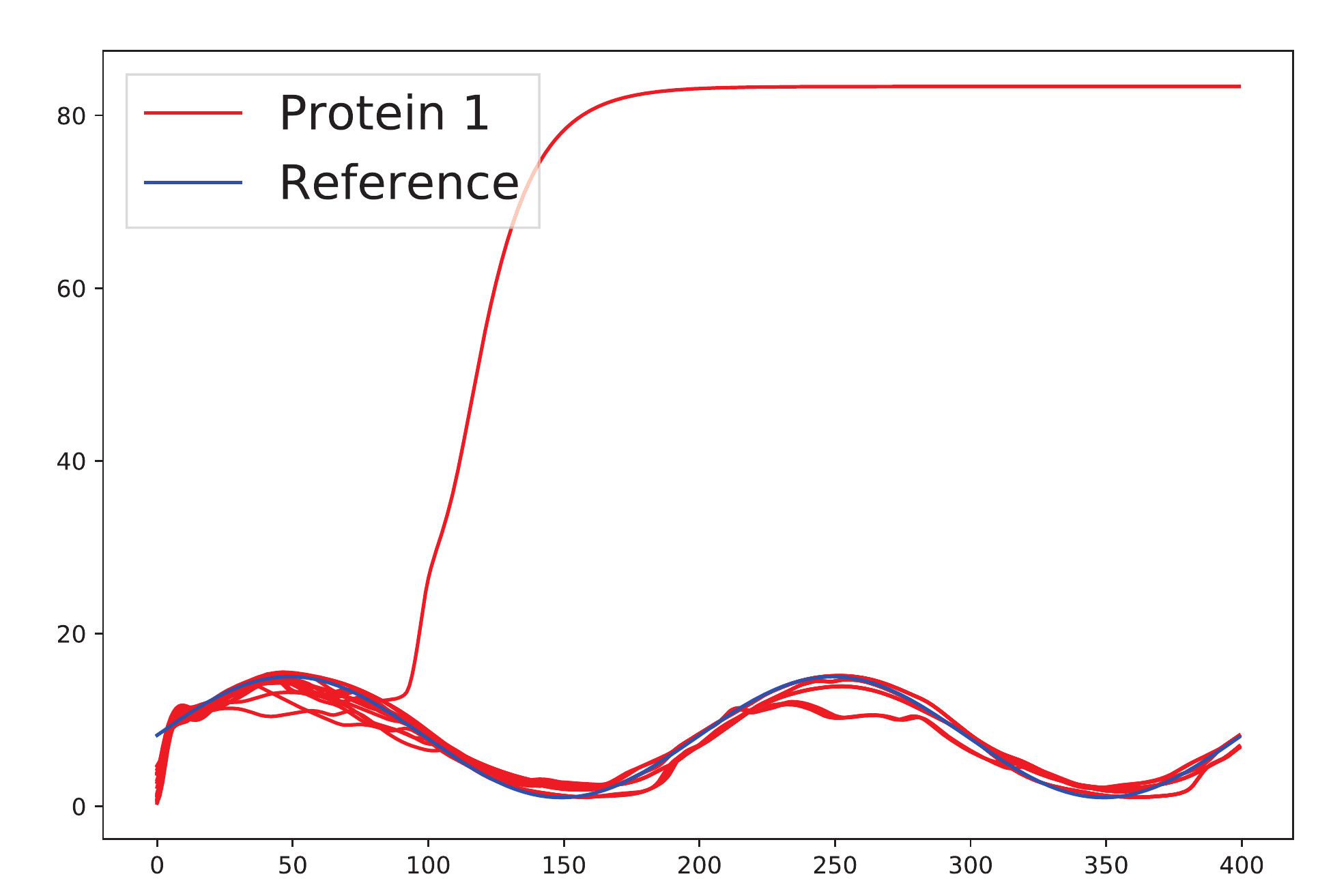}
    }
    \centering
    \subfigure[LAC-CompGRN]{
    \includegraphics[width=0.42\columnwidth]{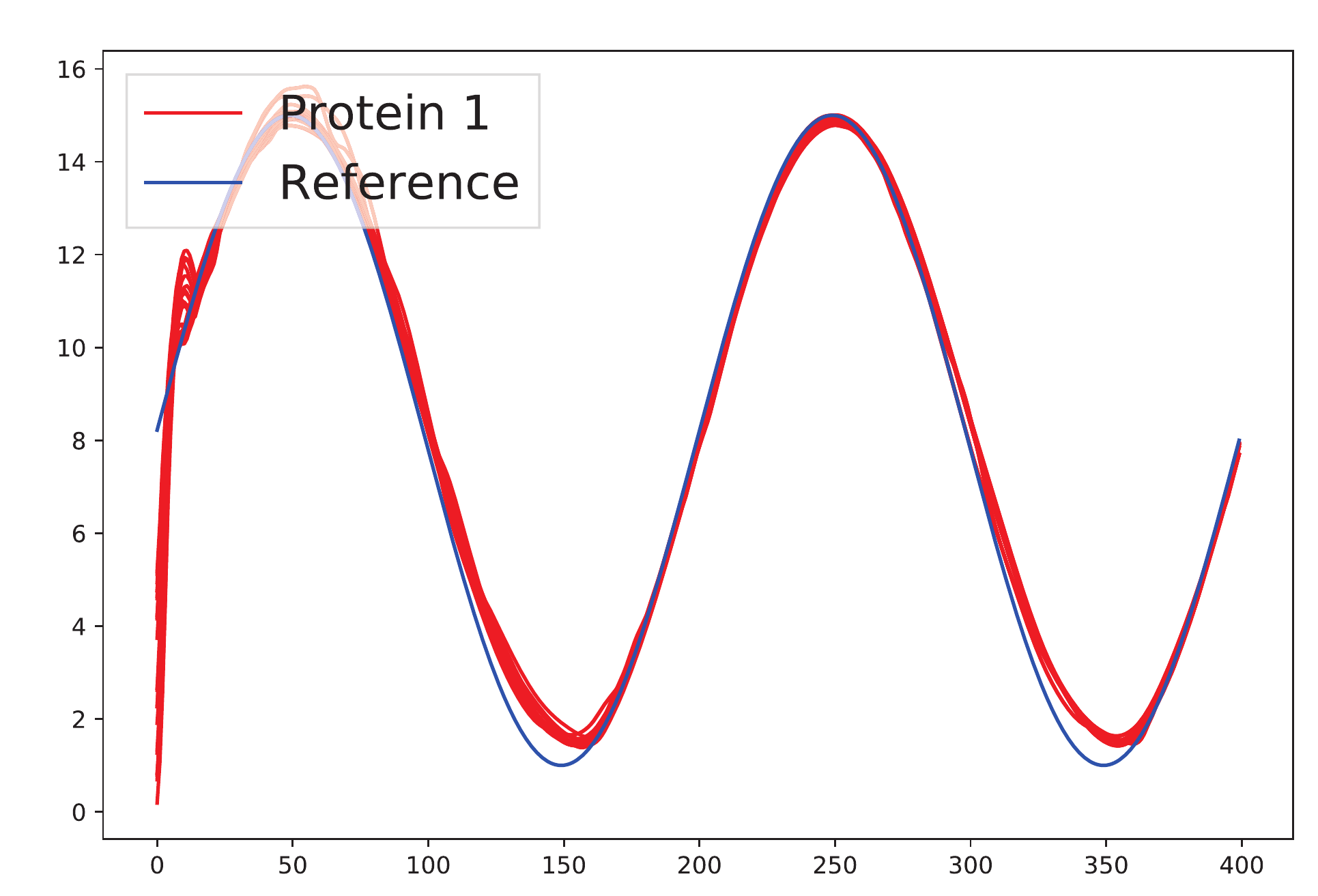}
    }
    \centering
    \subfigure[SAC-CompGRN]{
    \includegraphics[width=0.42\columnwidth]{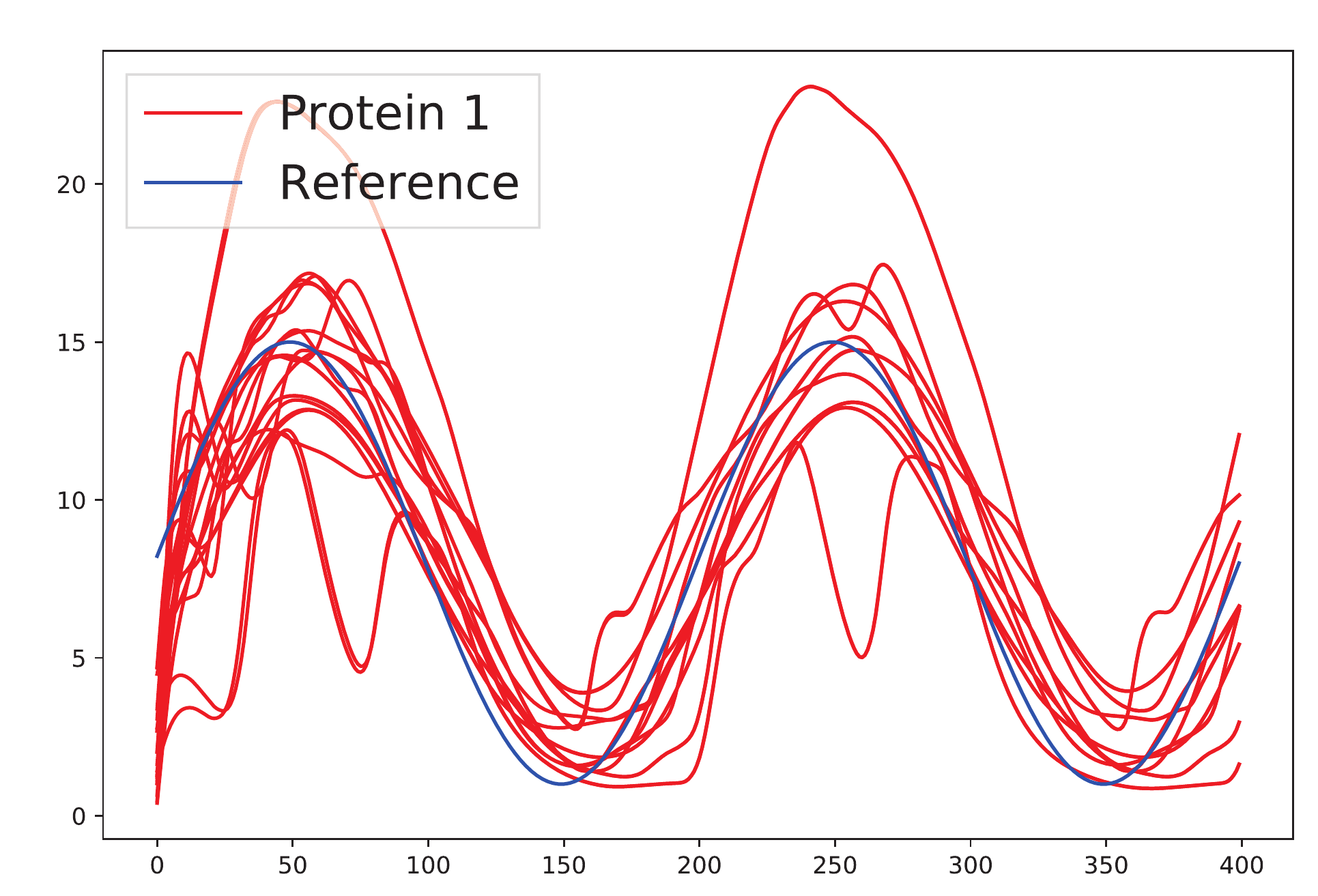}
    }
    \vspace{-0.15cm}
    \caption{State trajectories over time under policies trained by LAC and SAC in the GRN and CompGRN. In each experiment, the policies are tested over 20 random initial states and all the resulting trajectories are displayed above.
    The X-axis indicates the time and Y-axis shows the concentration of the target protein--- Protein 1.} 
    \label{fig:additional stability evaluation-1}
    \vspace{-0.2cm}
\end{figure}

We evaluated the trained policies in the GRN and CompGRN and the results are shown in \autoref{fig:additional stability evaluation-1}. In our experiments, we found that the LAC agents stabilize the systems well. All the state trajectories converge to the reference signal eventually (see \autoref{fig:additional stability evaluation-1} a and c). On the contrary, without stability guarantee, the state trajectories either diverge (see \autoref{fig:additional stability evaluation-1} b), or continuously oscillate around the reference trajectory (see \autoref{fig:additional stability evaluation-1} d).

\vspace{-0.3cm}

\subsection{Empirical Evaluation on Robustness and Generalization}\label{sec:exp:evaluation}
It is well-known that over-parameterized policies are prone to become overfitted to a specific training environment. The ability of generalization is the key to the successful implementation of RL agents in an uncertain real-world environment. In this part, we first evaluate the robustness of policies in the presence of parametric uncertainties and process noise. Then, we test the robustness of controllers against external disturbances. Finally, we evaluate whether the policy is generalizable by setting different reference signals. To make a fair comparison, we removed the policies that did not converge in SAC and only evaluate the ones that perform well during training. During testing, we found that SPPO appears to be prone to variations in the environment, thus the evaluation results are contained in Appendix \cite{han2020actor}.

\begin{figure}[htb]
    \centering
    \subfigure[LAC-CartPole]{
    \includegraphics[width=0.45\columnwidth]{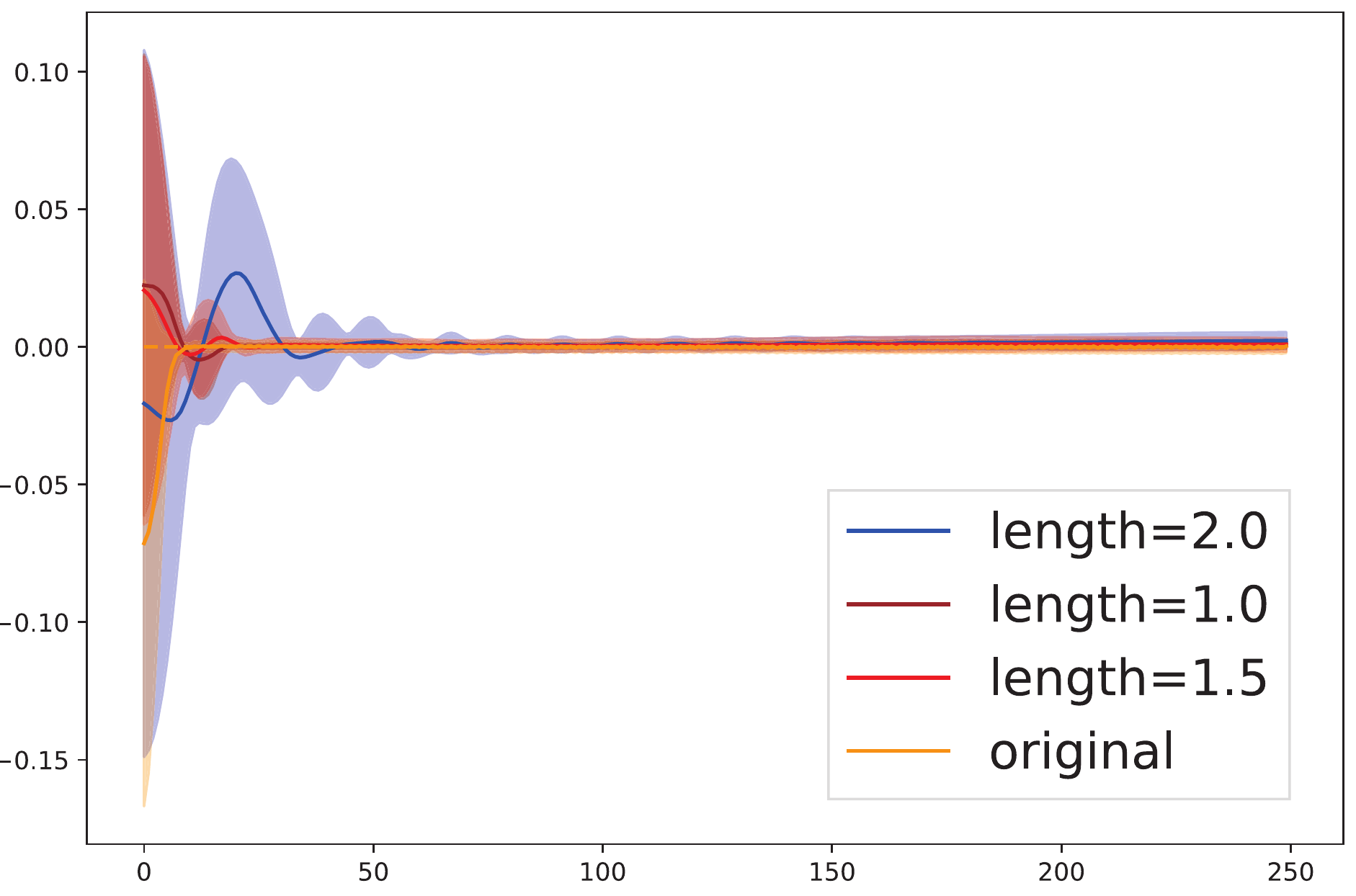}
    \label{LAC-CartPole}
    }
    \centering
    \subfigure[SAC-CartPole]{
    \includegraphics[width=0.45\columnwidth]{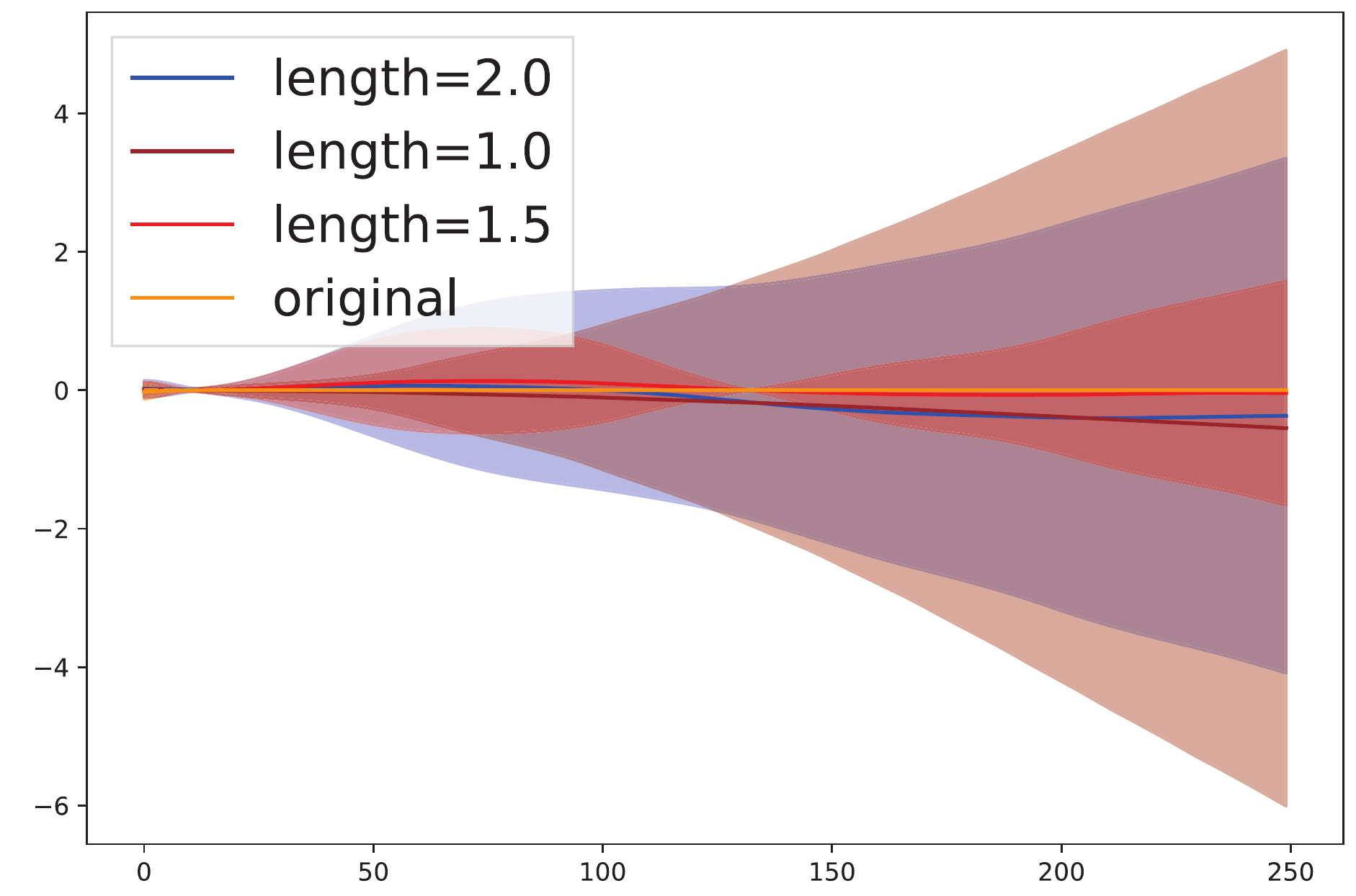}
    \label{SAC-CartPole}
    }
    \centering
    \subfigure[LAC-GRN]{
    \includegraphics[width=0.45\columnwidth]{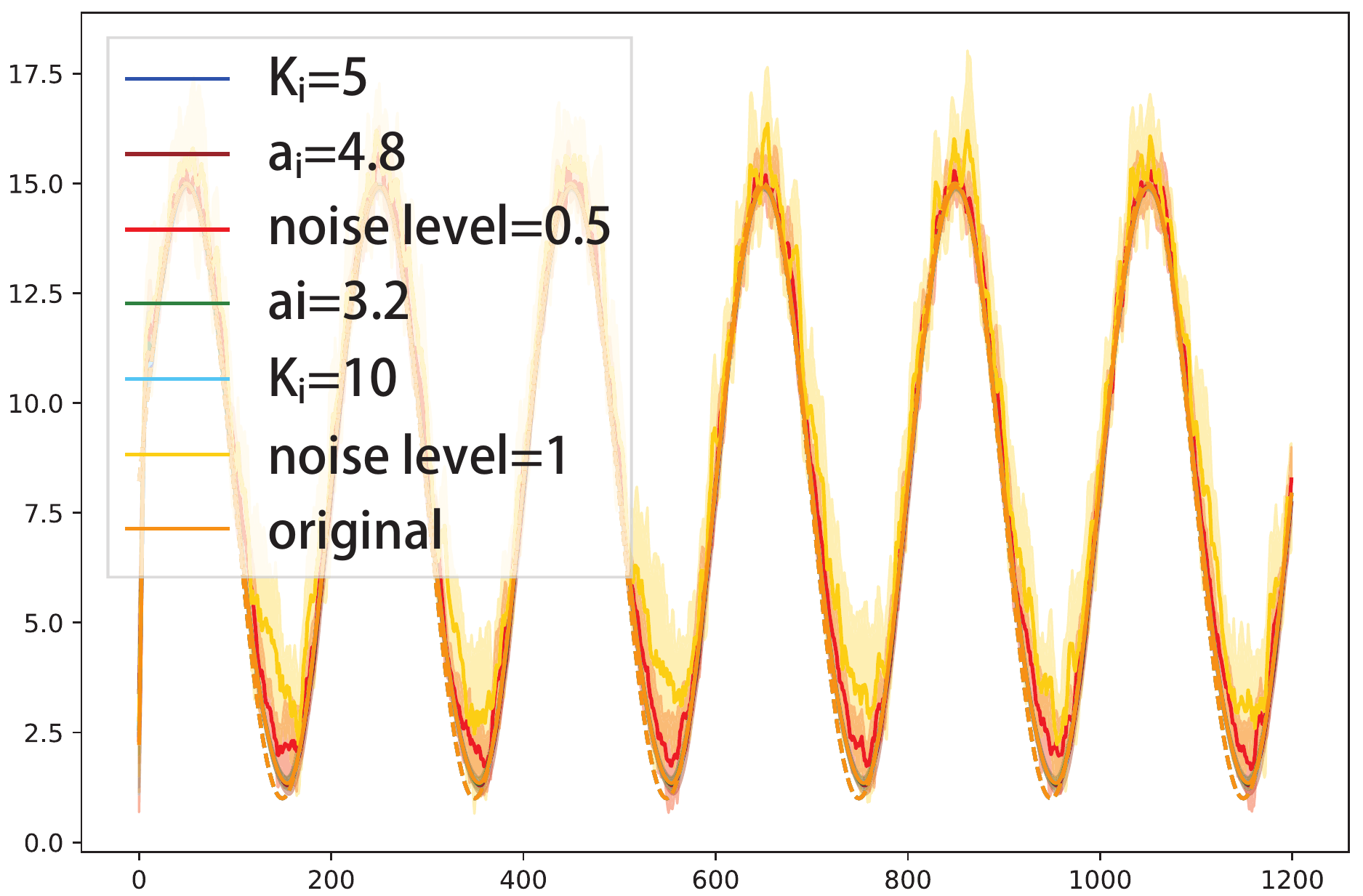}
    \label{LAC-GRN}
    }
    \centering
    \subfigure[SAC-GRN]{
    \includegraphics[width=0.45\columnwidth]{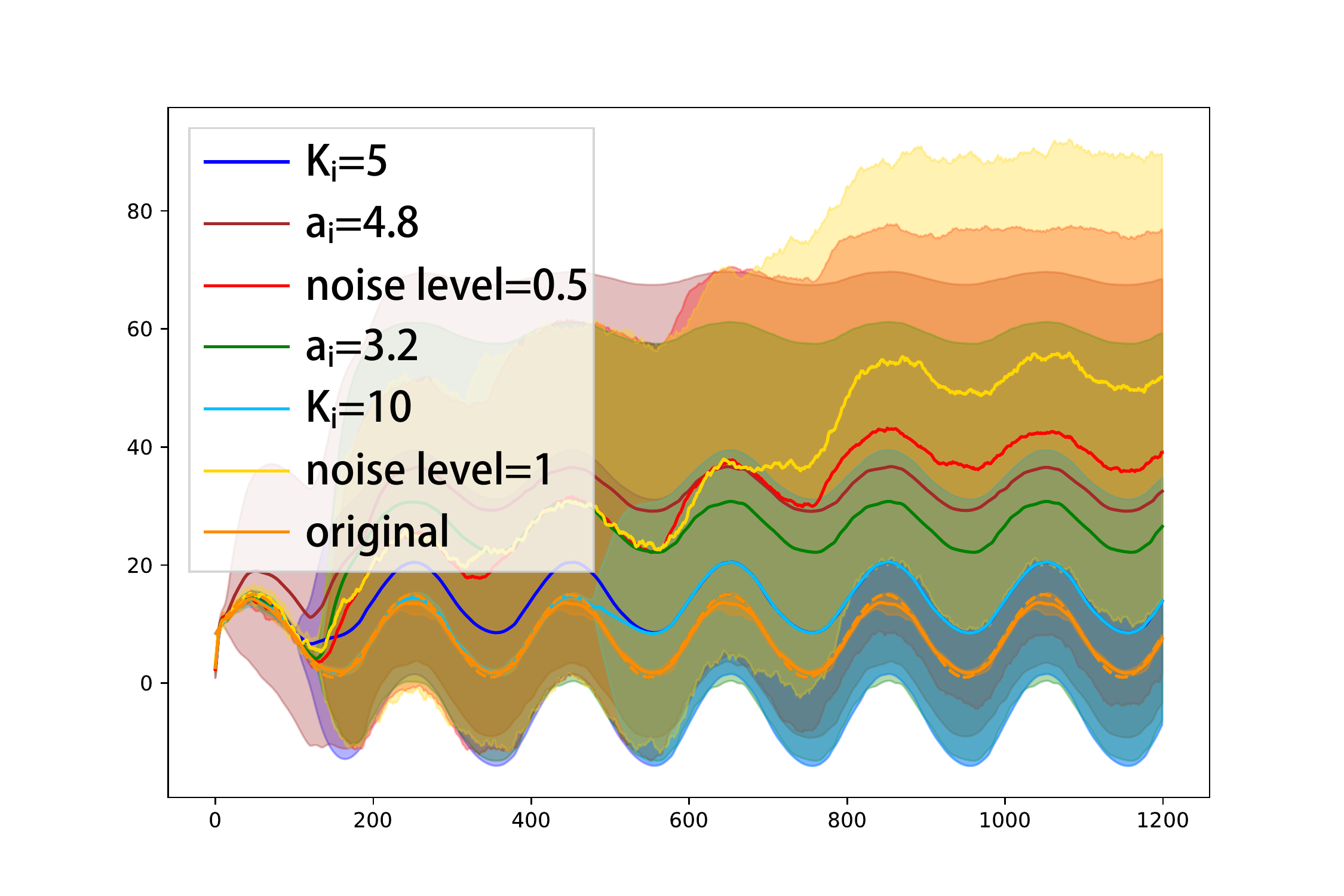}
    \label{SAC-GRN}
    }
    \caption{LAC and SAC agents in the presence of dynamic uncertainties. The solid line indicates the average trajectory and shadowed region for the 1-SD confidence interval. In (a) and (b), the pole length is varied during the inference. In (c) and (d), three parameters are selected to reflect the uncertainties in gene expression. 
    The X-axis indicates the time and Y-axis shows the angle of the pole in (a,b) and concentration of target protein in (c,d), respectively. The dashed line indicates the reference signal. The line in orange indicates the dynamic in the original environment. For each curve, only the noted parameter is different from the original setting.}
    \label{fig:uncertainty}
    \vspace{-0.2cm}
\end{figure}

\subsubsection{Robustness to dynamic uncertainties} \label{sec:Robustness to dynamic uncertainty}
In this part, during the inference, we vary the system parameters and introduce process noises in the model/simulator to evaluate the algorithm's robustness. In CartPole, we vary the length of pole $l$. In GRN, we vary the promoter strength $a_i$ and dissociation rate $K_i$. Due to stochastic nature in gene expression, we also introduce uniformly distributed noise ranging from $[-\delta,\delta]$ (we indicate the \emph{noise level} by $\delta$) to the dynamic of GRN. The state trajectories of the closed-loop system under LAC and SAC agents in the varied environment are demonstrated in \autoref{fig:uncertainty}.

As shown in Figure~\ref{fig:uncertainty} (a, c), the policies trained by LAC are very robust to the dynamic uncertainties and achieve high tracking precision in each case. On the other hand, though SAC performs well in the original environment (Figure~\ref{fig:uncertainty} (b, d)), it fails in all of the varied environments.

\subsubsection{Robustness to disturbances}
\begin{figure}[htb] 
\vspace{-0.1cm}
\centering
\subfigure[CartPole]{
\includegraphics[width=0.45\columnwidth]{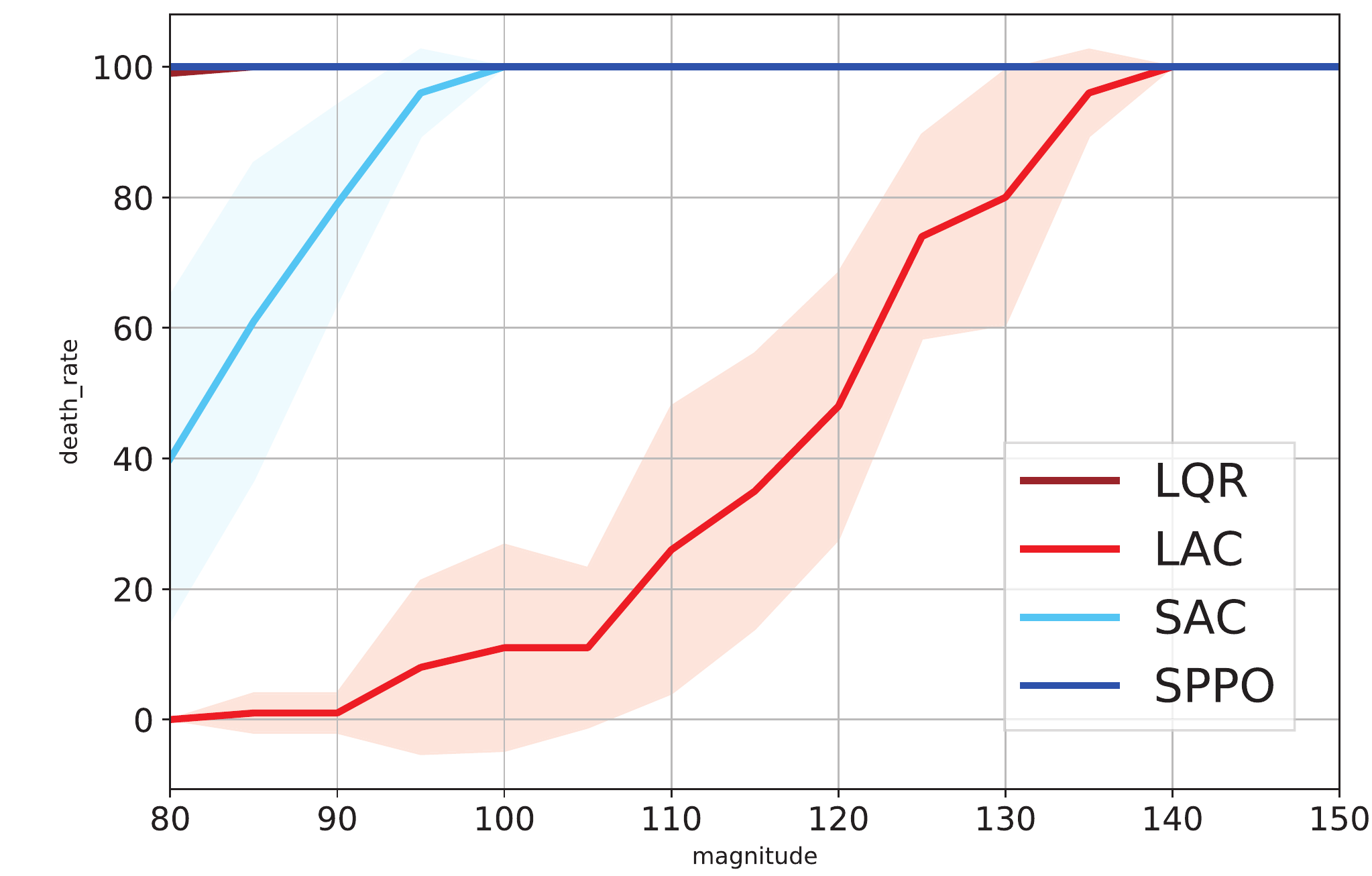}
}
\centering
\subfigure[HalfCheetah]{
\includegraphics[width=0.45\columnwidth]{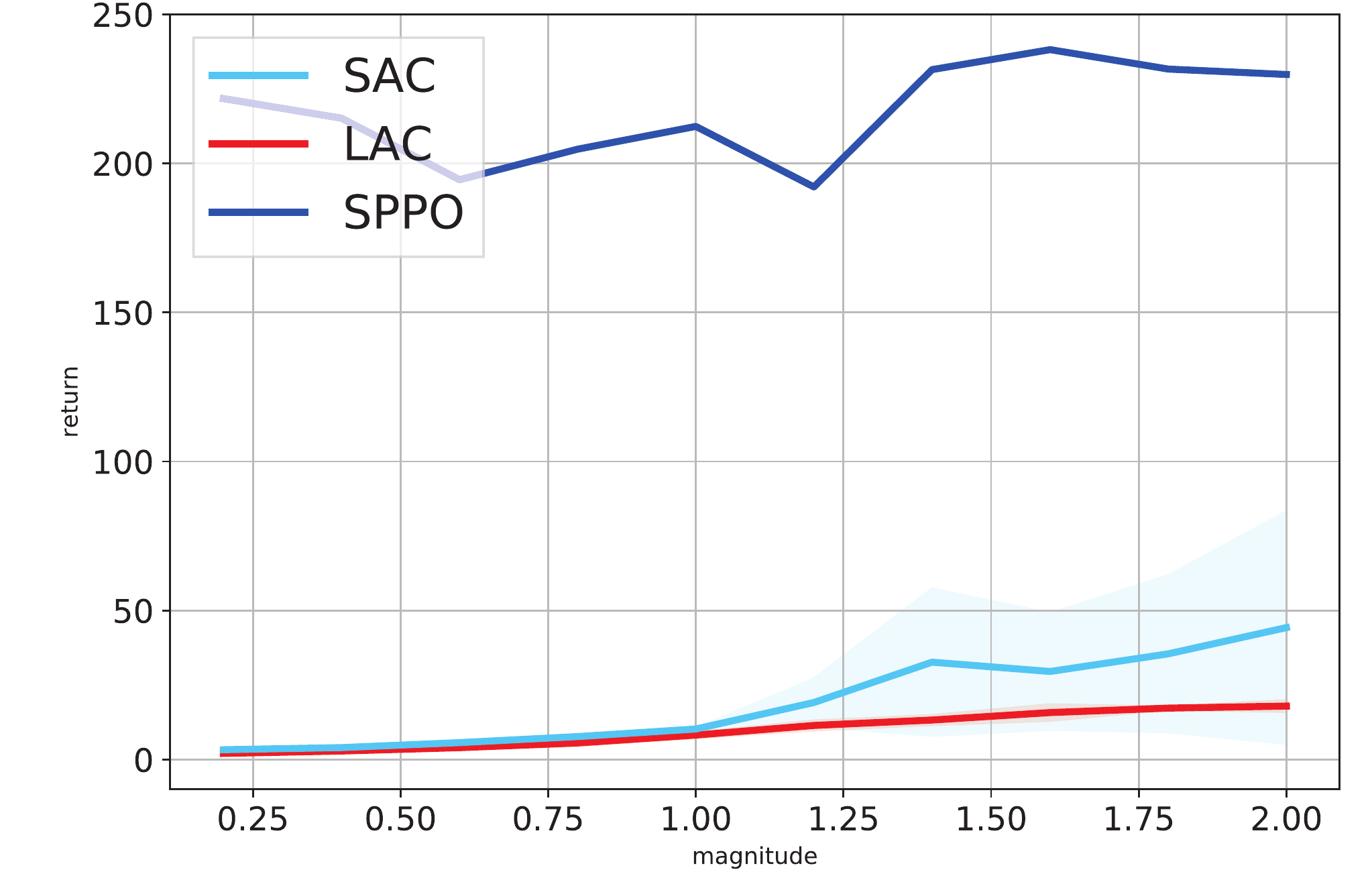}
}
\centering
\subfigure[FetchReach]{
\includegraphics[width=0.45\columnwidth]{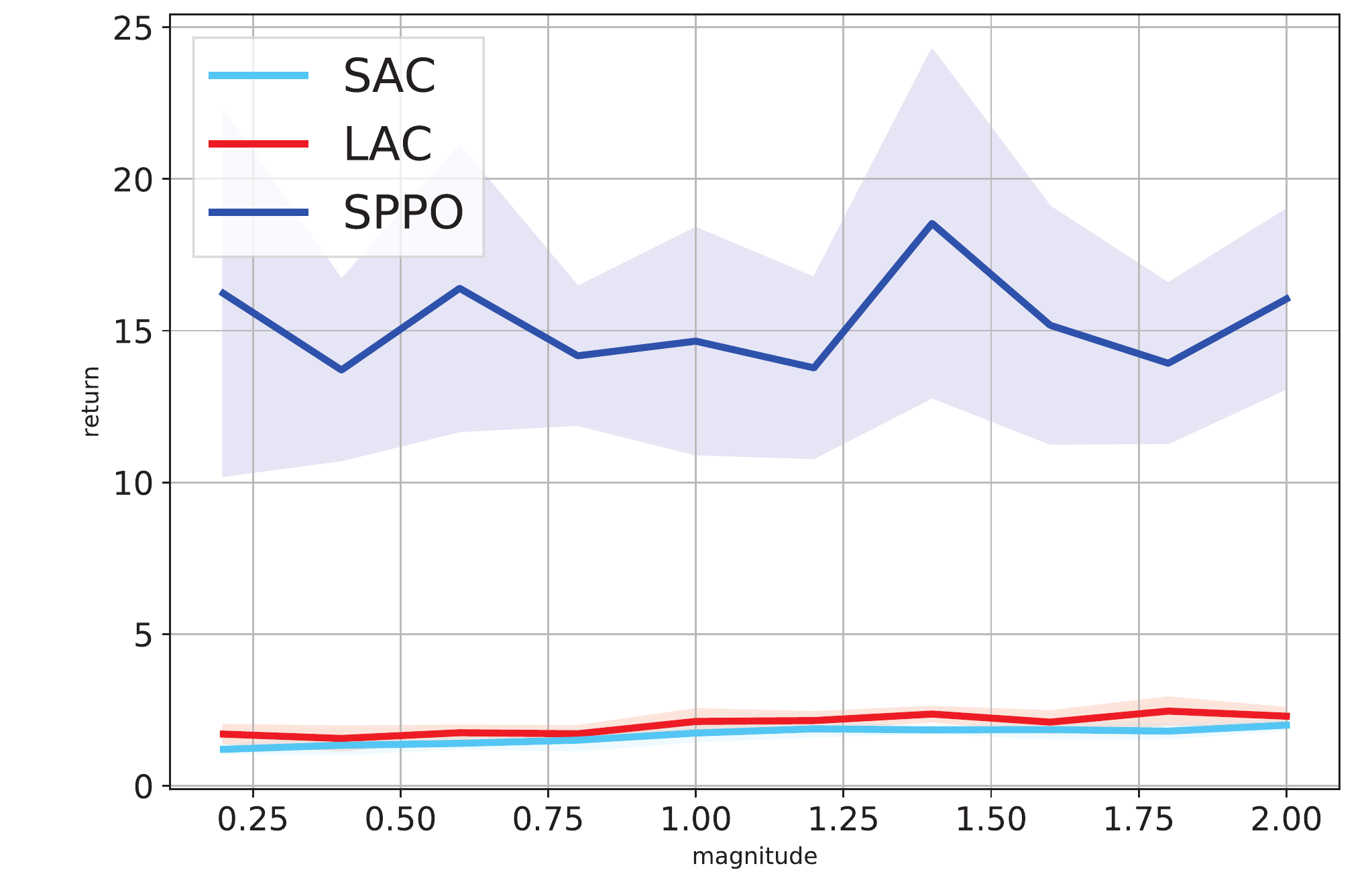}
}
\centering
\subfigure[GRN]{
\includegraphics[width=0.45\columnwidth]{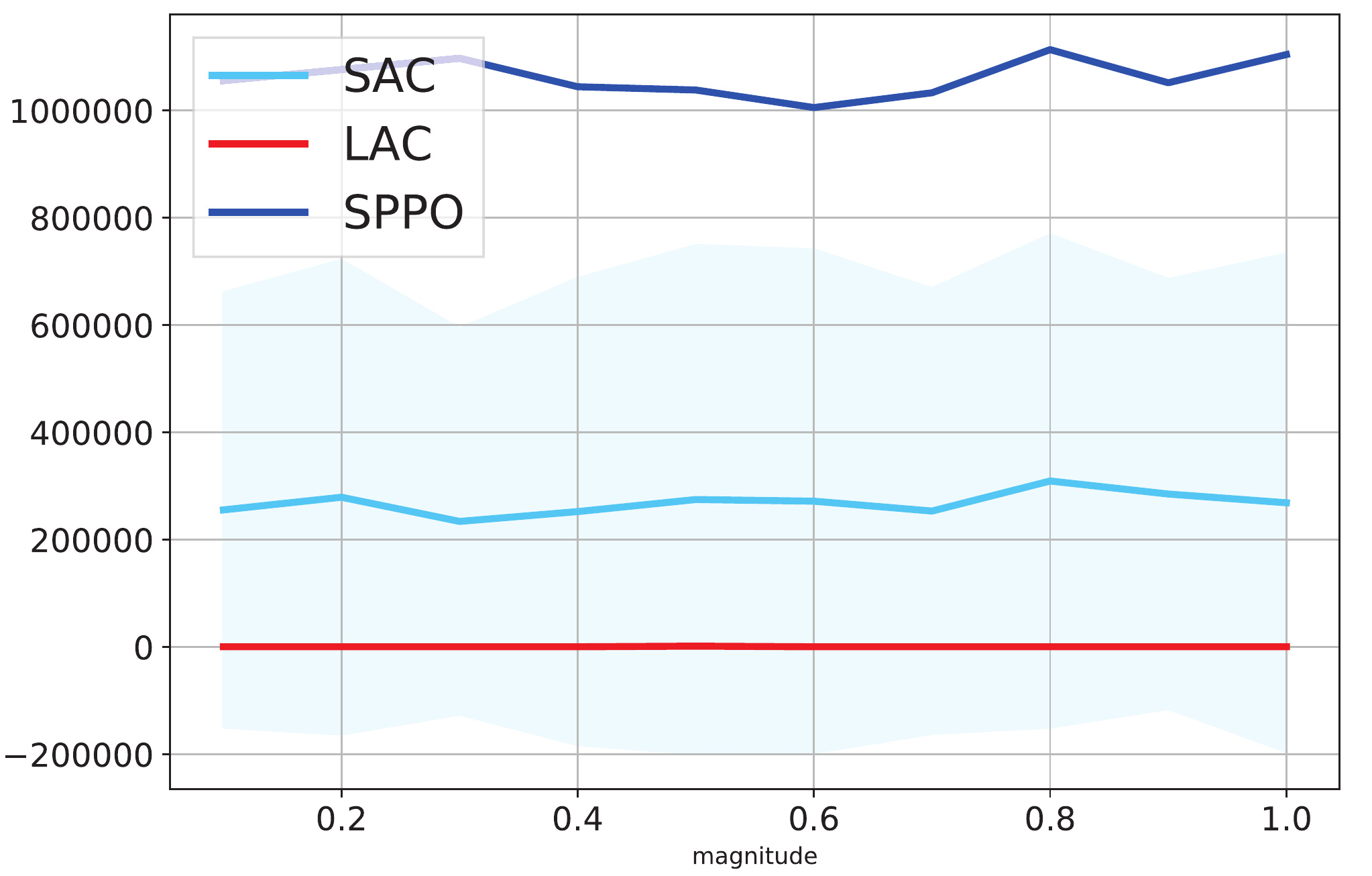}
}
\centering
\subfigure[Swimmer]{
\includegraphics[width=0.46\columnwidth]{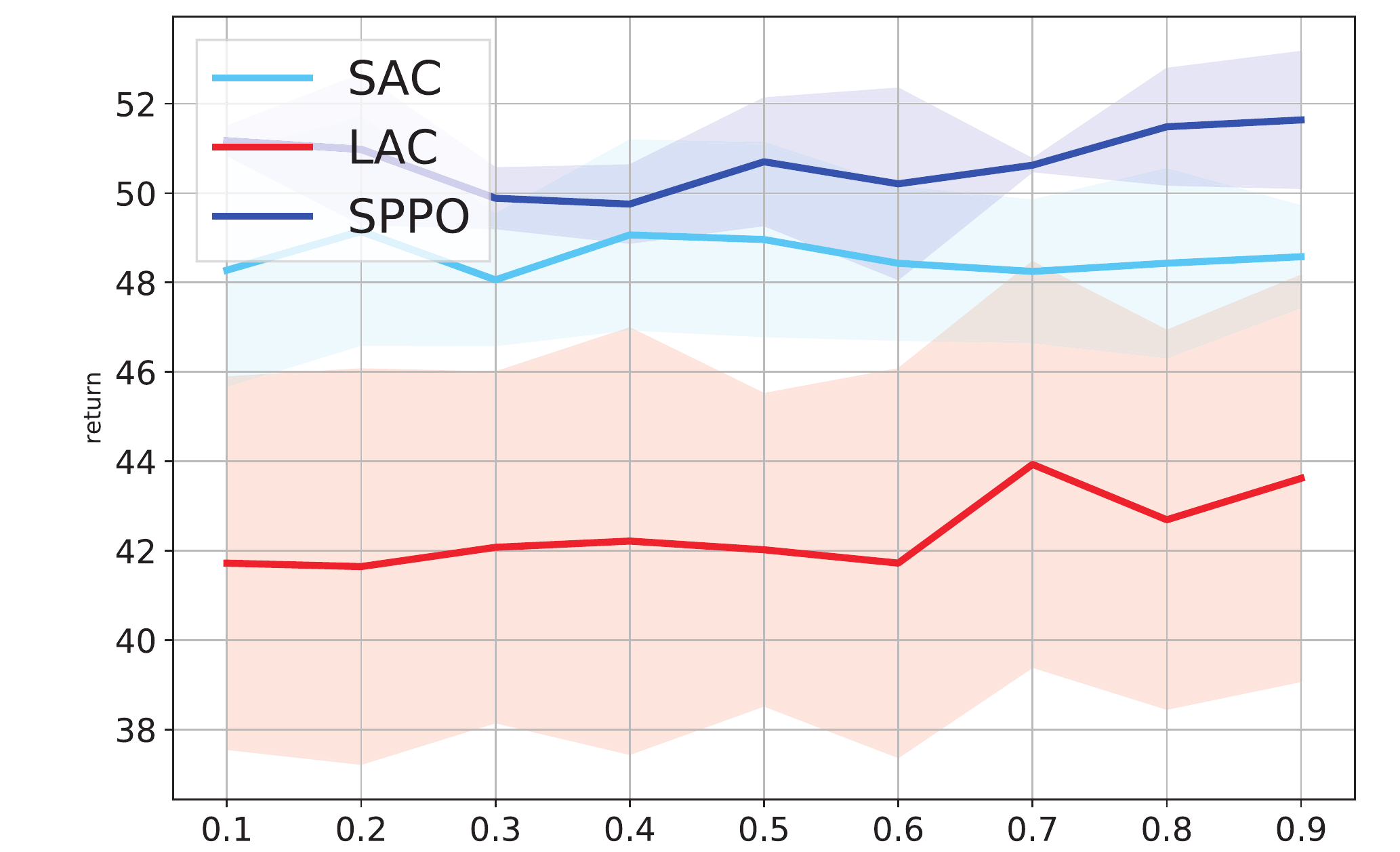}
}
\centering
\subfigure[Minitaur]{
\includegraphics[width=0.46\columnwidth]{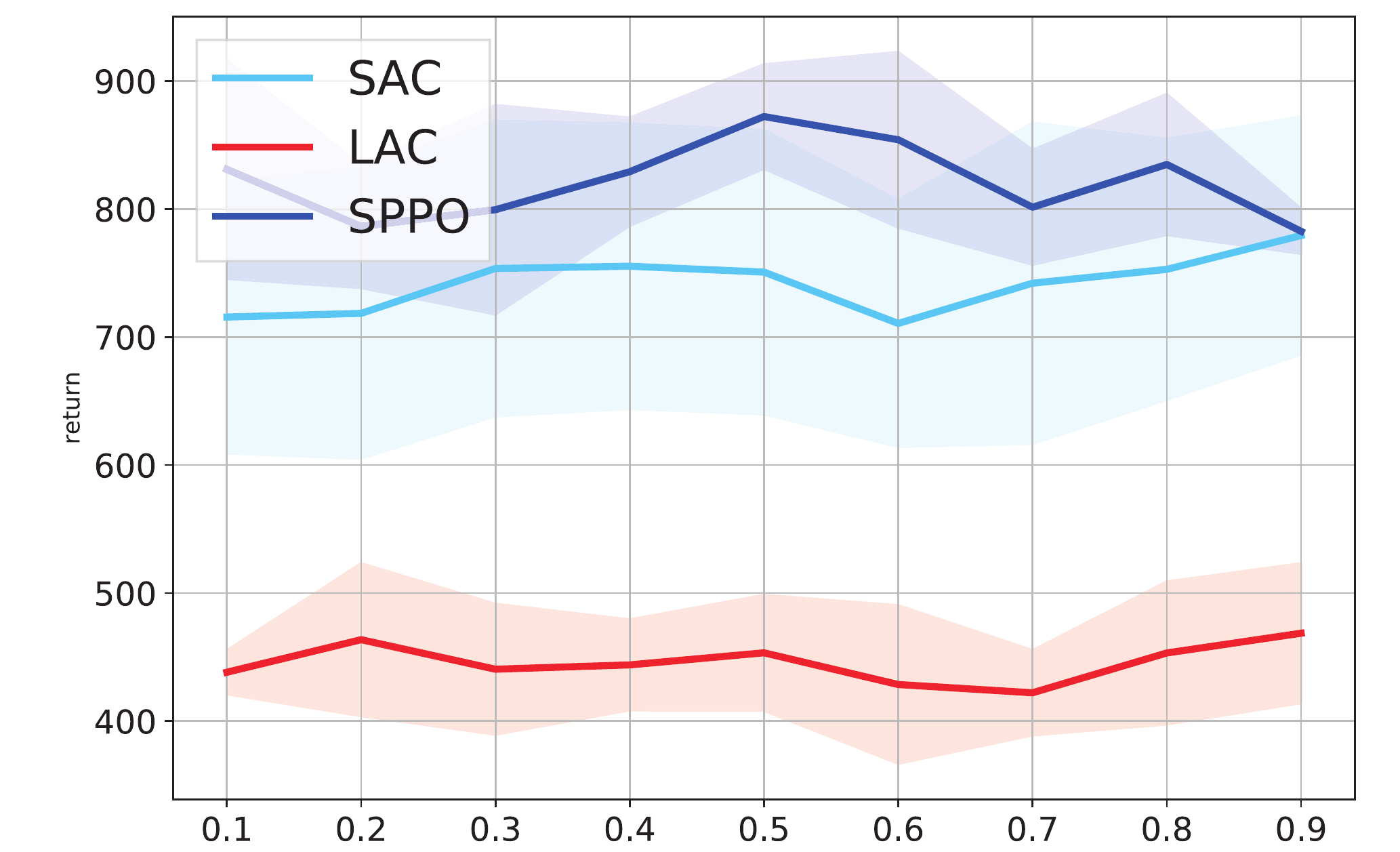}
}
\vspace{-0.2cm}
\caption{Performance of LAC, SAC, SPPO and LQR in the presence of persistent disturbances with different magnitudes. The X-axis indicates the magnitude of the applied disturbance. The Y-axis indicates the death rate in CartPole (a) and the cumulative cost in other examples (b)-(d). All of the trained policies are evaluated for 100 trials in each setting.}
\label{fig:death rate}
\vspace{-0.15cm}
\end{figure}

An inherent property of a stable system is to recover from perturbations such as external forces and wind. To show this, we introduce periodic external disturbances with different magnitudes in each environment and observe the performance difference between policies trained by LAC and SAC. We also include LQR as the model-based baseline. In CartPole, the agent may fall over when interfered by an external force, ending the episode in advance. Thus in this task, we measure the robustness of controllers through the death-rate, i.e., the probability of falling over after being disturbed. For other tasks where the episodes are always of the same length and we measure the robustness of controller by the variation in the cumulative cost. Under each disturbance magnitude, the policies are tested for $100$ trials and the performance is shown in \autoref{fig:death rate}.

As shown in \autoref{fig:death rate}, the controllers trained by LAC outperform SAC and LQR to a great extent in CartPole and GRN (lower death rate and cumulative cost). In HalfCheetah, SAC and LAC are both robust to small external disturbances while LAC is more reliable to larger ones. In FetchReach, SAC and LAC perform reliably across all of the external disturbances. The difference between SAC and LAC becomes obvious in GRN, Swimmer, and Minitaur, where the dynamics are more vulnerable to the external disturbances.
In all of the experiments, SPPO agents could hardly sustain any external disturbances.

\vspace{-0.1cm}
\begin{figure}[htb]
\vspace{-0.2cm}
    \centering
    \subfigure[LAC]{
    \includegraphics[width=0.45\columnwidth]{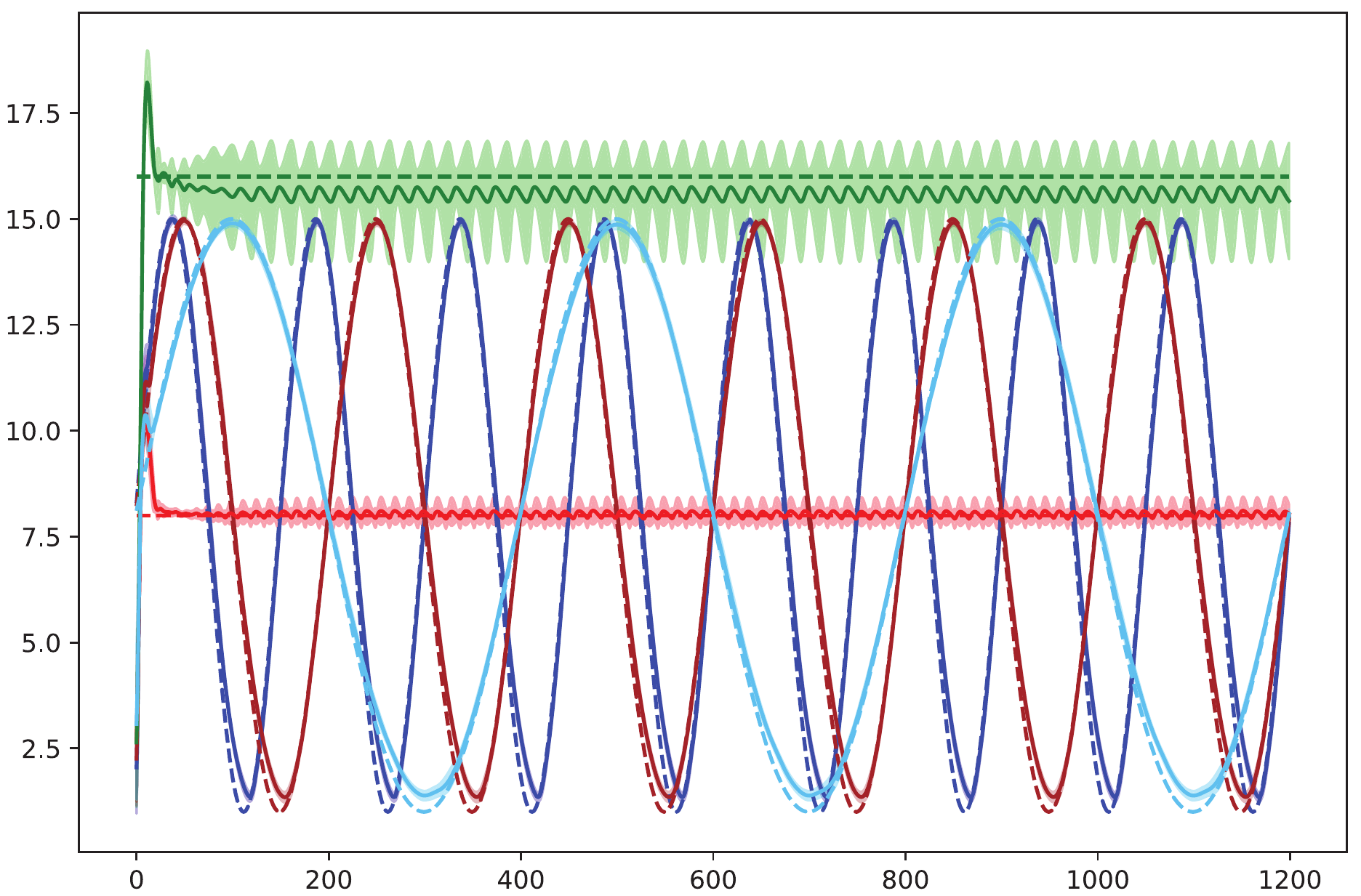}
    }
    \centering
    \subfigure[SAC]{
    \includegraphics[width=0.45\columnwidth]{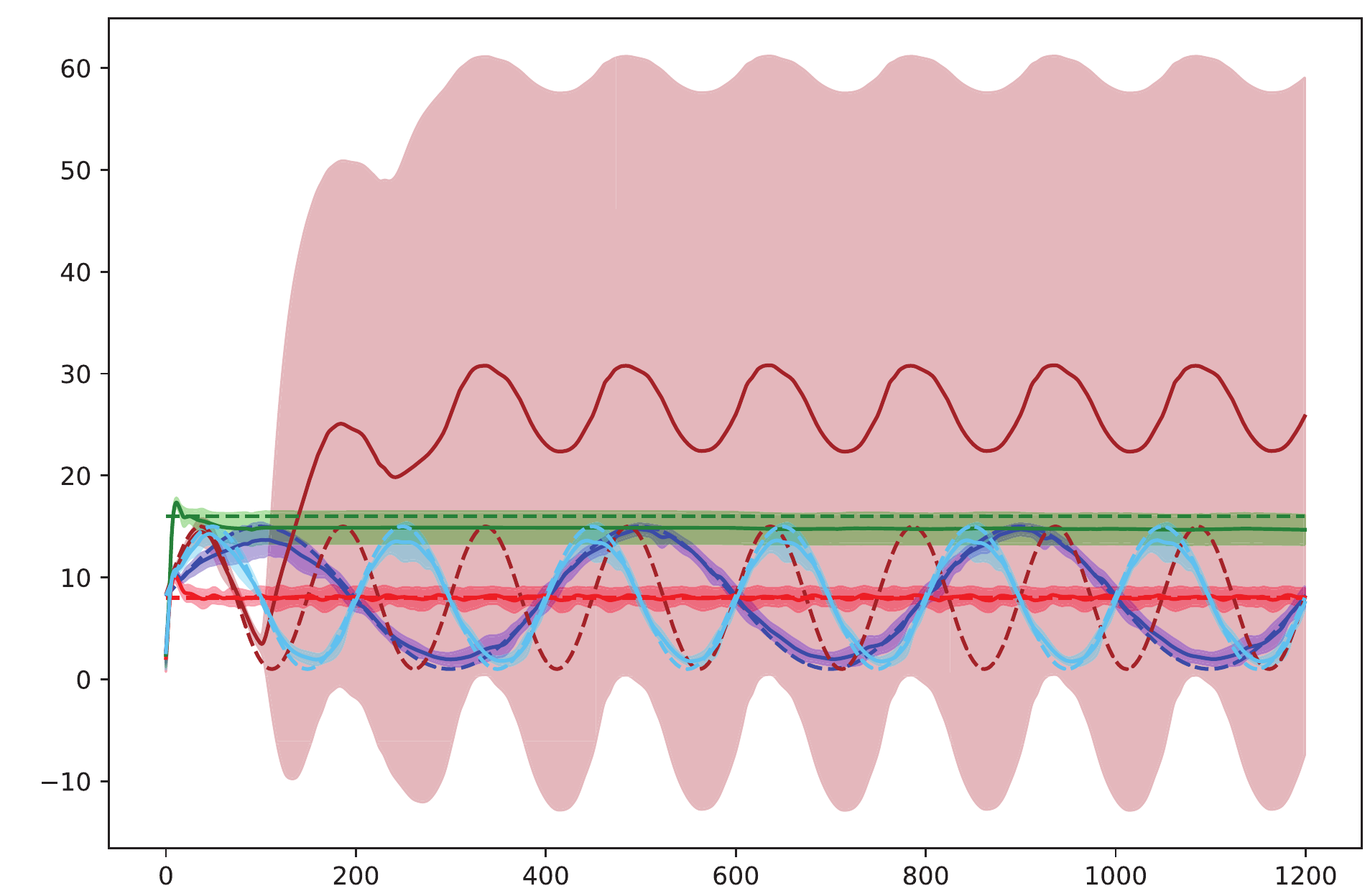}
    }
    \caption{State trajectories under policies trained by LAC and SAC when tracking different reference signals. The solid line indicates the average trajectory and shadowed region for the 1-SD confidence interval. The X-axis indicates the time and Y-axis shows the concentration of protein to be controlled. Dashed lines in different colors are the different reference signals: sinusoid with a period of 150 (brown); sinusoid with a period of 200 (sky-blue); sinusoid with a period of 400 (blue); constant reference of 8 (red); constant reference of 16 (green).}
    \label{fig:different reference}
    \vspace{-0.3cm}
\end{figure}

\vspace{-0.1in}
\subsubsection{Generalization over different tracking references}\label{sec:Generalization over different tracking references}
In this part, we introduce four different reference signals that are unseen during training in the GRN: sinusoids with periods of 150 (brown) and 400 (blue), and the constant reference of 8 (red) and 16 (green). We also show the original reference signal used for training (skyblue) as a benchmark. Reference signals are indicated in \autoref{fig:different reference} by the dashed line in respective colors. All of the trained policies are tested for 10 times with each reference signal. The average dynamics of the target protein are shown in \autoref{fig:different reference} with the solid line, while the variance of dynamic is indicated by the shadowed area.

As shown in \autoref{fig:different reference}, the policies trained by LAC could generalize well to follow previously unseen reference signals with low variance (dynamics are very close to the dashed lines), regardless of whether they share the same mathematical form with the one used for training. On the other hand, though SAC tracks the original reference signal well after the training trials without convergence being removed (see the sky-blue lines), it is still unable to follow some of the reference signals (see the brown line) and possesses larger variance than LAC.

\subsection{Influence of Different Lyapunov candidates} \label{sec:exp:candidates}

\begin{figure}[htb]
\centering
\subfigure[Horizon-Training]{
\includegraphics[width=0.45\columnwidth]{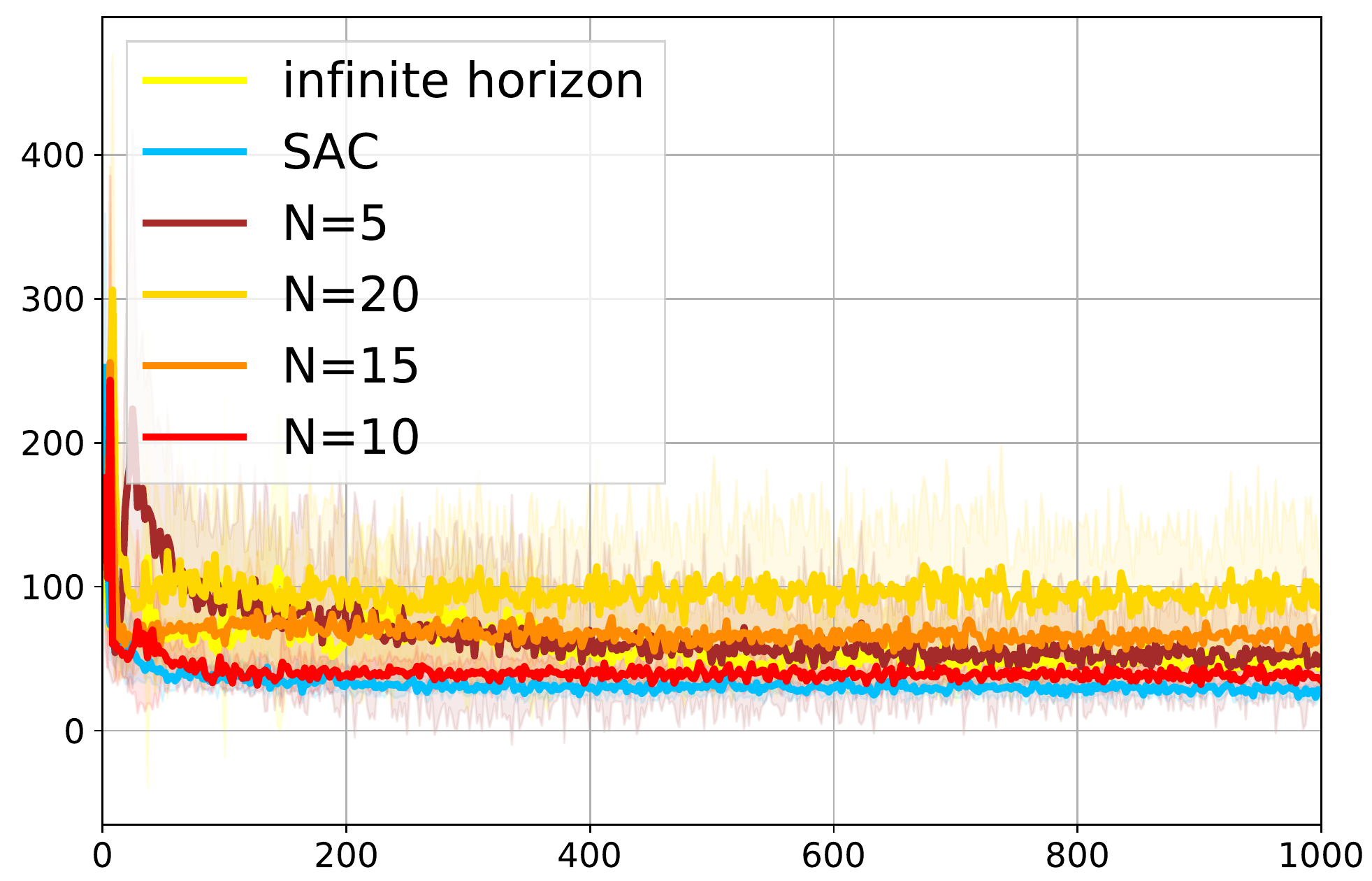}
}
\centering
\subfigure[Horizon-Robustness]{
\includegraphics[width=0.45\columnwidth]{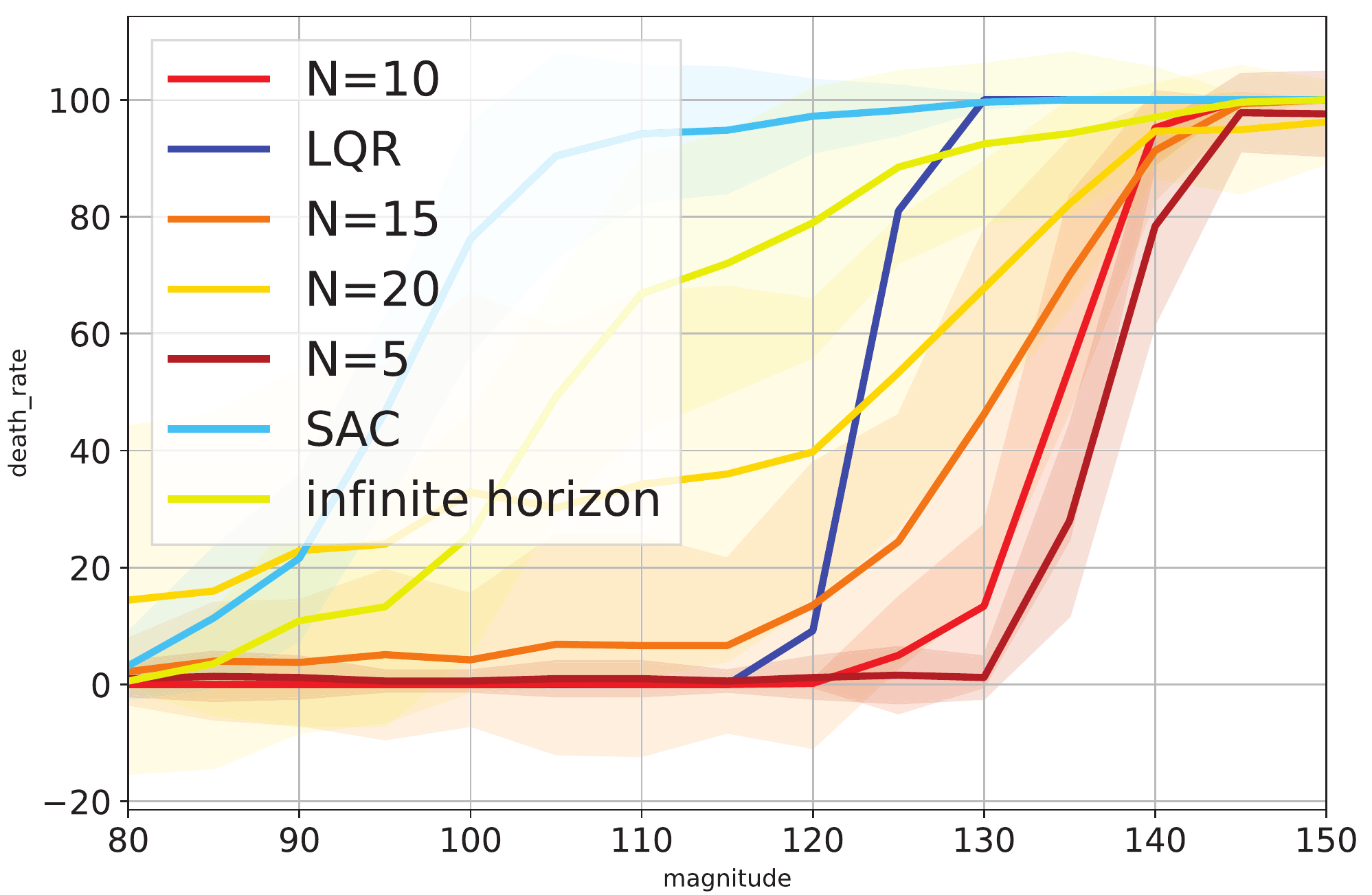}
}
\caption{Influence of different Lyapunov candidates. In (a), the Y-axis indicates cumulative cost during training and the X-axis indicates the total time steps in thousand. (b) shows the death-rate of policies in the presence of instant impulsive force $F$ ranging from 80 to 150 Newton.}
\label{fig:different horizons}
\vspace{-0.15cm}
\end{figure}

As an independent interest, we evaluate the influence of choosing different Lyapunov candidates in this part. First, we adopt candidates of different time horizon $N\in\{5,10,15,20,\infty\}$ to train policies in the CartPole example and compare their performance in terms of cumulative cost and robustness. Here, $N=\infty$ implies using value function as the Lyapunov candidate. Both of the Lyapunov critics are parameterized as \eqref{eq:lyapunov critic parameterization}. For evaluation of robustness, we apply an impulsive force $F$ at $100_{\text{th}}$ instant and observe the death-rate of trained policies. The results are demonstrated in \autoref{fig:different horizons}.

\vspace{-0.3cm}

As shown in \autoref{fig:different horizons}, both choices of Lyapunov candidates converge fast and achieve comparable cumulative cost at convergence. However, in terms of robustness, the choice of $N$ plays an important role. As observed in \autoref{fig:different horizons} (b), the robustness of the controller decreases as the time horizon $N$ increases. Besides, it is interesting that LQR is more robust than SAC when faced with instant impulsive disturbance.

\section{CONCLUSIONS and DISCUSSIONS}
In this paper, we proposed a data-based approach for analyzing the stability of discrete-time nonlinear stochastic systems modeled by Markov decision process, by using the classic Lyapunov's method in control theory. By employing the stability condition as a critic, an actor-critic algorithm is proposed to learn a controller/policy to ensure the closed-loop stability in stabilization and tracking tasks. We evaluated the proposed method in various examples and show that our method achieves not only comparable or superior performance compared with the state-of-the-art RL algorithm but also outperforms impressively in terms of robustness to uncertainties such as model parameter variations and external disturbances. For future work, it might be interesting to extend this method to constrained Markov decision process in which state and action constraints are considered. Also, to quantify the robustness induced by the stability will be investigated.

\newpage
\clearpage
\onecolumn


\begin{center}
	{\Large \textbf{Appendix}}
\end{center}

\makeatletter
\renewcommand{\thesection}{S\arabic{section}}   
\renewcommand{\thesubsection}{S\arabic{subsection}}   
\renewcommand{\thetable}{S\arabic{table}}   
\renewcommand{\thefigure}{S\arabic{figure}}

\setcounter{section}{0}

\section{Experiment Setup}\label{app:Experiment Setup}

\begin{figure}[h]
    \centering
    \subfigure[Cartpole]{
    \includegraphics[width=0.17\columnwidth]{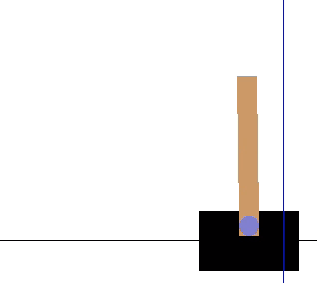}
    }
    \centering
    \subfigure[HalfCheetah]{
    \includegraphics[width=0.15\columnwidth]{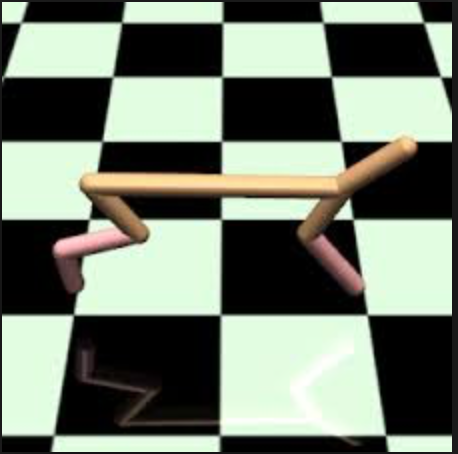}
    }
    \centering
    \subfigure[FetchReach]{
    \includegraphics[width=0.18\columnwidth]{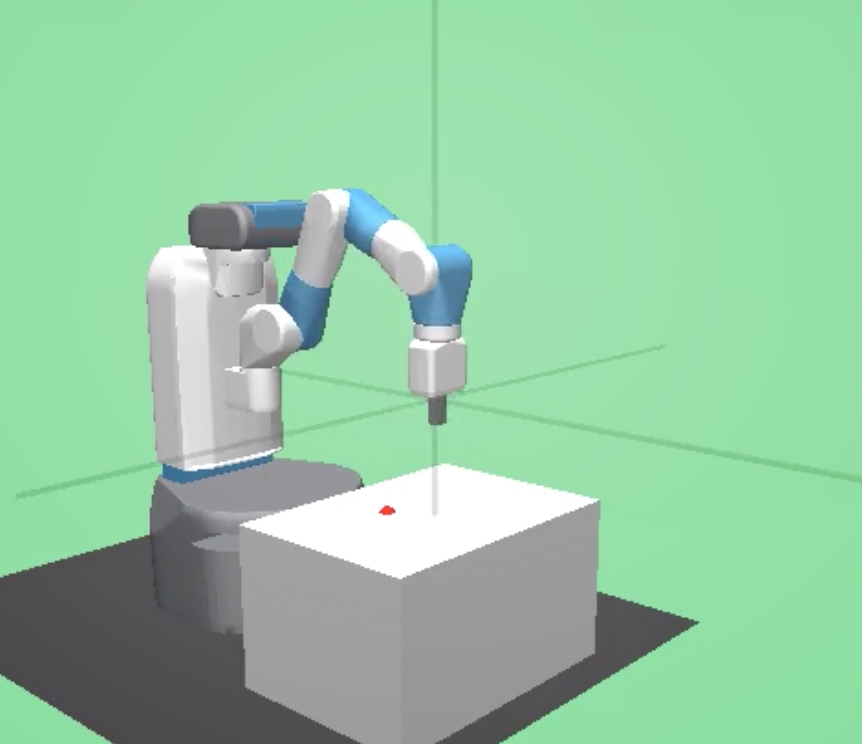}
    }
    \centering
    \subfigure[Swimmer]{
    \includegraphics[width=0.15\columnwidth]{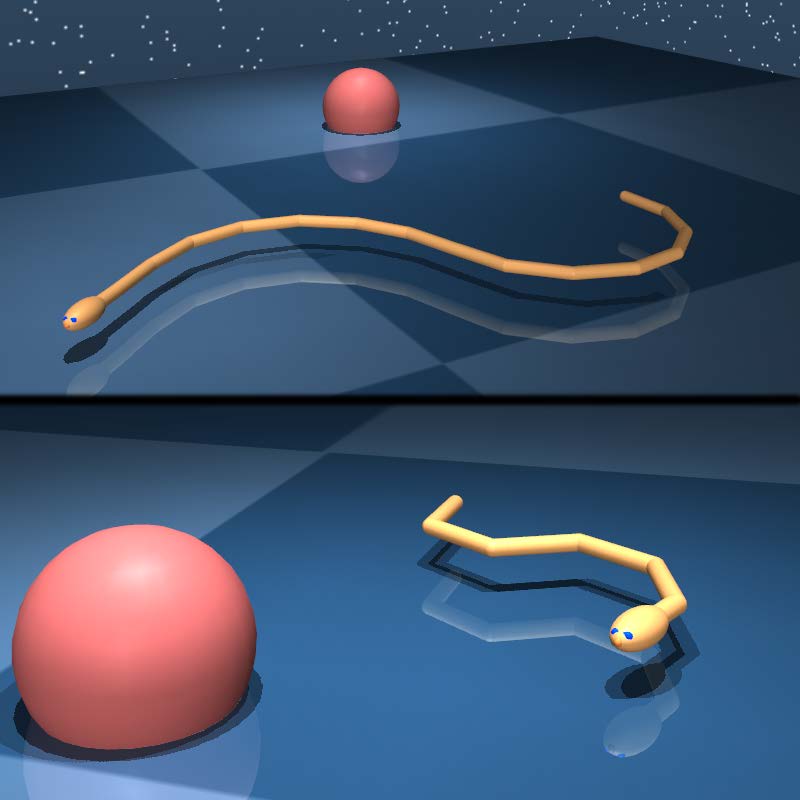}
    }
    \centering
    \subfigure[Minitaur]{
    \includegraphics[width=0.2\columnwidth]{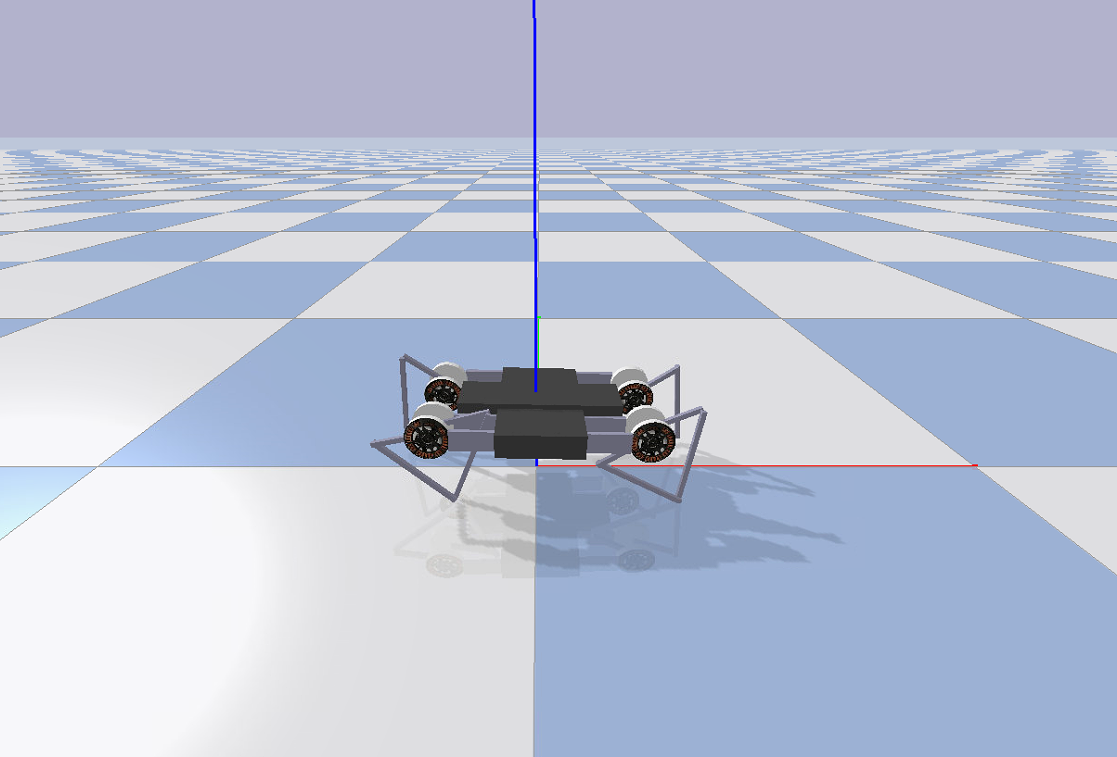}
    }
    \caption{Snapshot of environments using OpenAI Gym.}
    \label{fig:experiments_setup}
\end{figure}

We set up the experiment using OpenAi Gym \cite{brockman2016openai}, DeepMind Control Suite \cite{tassa2018deepmind} and PyBullet physics simulation platforms \cite{coumans2016pybullet}. A snapshot of the adopted environments in this paper can be found in Figure~\ref{fig:experiments_setup}.

\subsection{CartPole}

In this experiment, the controller is expected to maintain the pole vertically at a target position $x=0$. This is a modified version of CartPole in~\cite{brockman2016openai} with continuous action space. The action is the horizontal force applied upon the cart ($a\in[-20, 20]$).
 $x_{\text{threshold}}$ and $\theta_{\text{threshold}}$ represents the maximum of position and angle, respectively, $x_{\text{threshold}}=10$ and $\theta_{\text{threshold}}=20^\circ$. The episode ends if $\vert x\vert> x_{\text{threshold}}$ or $\vert \theta\vert> \theta_{\text{threshold}}$ and the episodes end in advance. Cost function $ r=(\frac{x}{ x_{\text{threshold}}})^2+20 *(\frac{\theta}{ \theta_{\text{threshold}}})^2 $. The episodes are of length 250.
For robustness evaluation in Section~\ref{sec:exp:evaluation}, we apply an impulsive disturbance force $F$ on the cart every 20 seconds, of which the magnitude ranges from 80 to 150 and the direction is opposite to the direction of control input. In Section~\ref{sec:exp:candidates}, the impulsive disturbance has the same magnitude range and direction with that in Section~\ref{sec:exp:evaluation}, but only applied once at instant $t=100$.

\subsection{HalfCheetah}
HalfCheetah is a modified version of that in Gym's robotics environment~\cite{brockman2016openai}. The task is to control a HalfCheetah (a 2-legged simulated robot) to run at the speed of $1$ $m/s$.
The reward is $r = (v-1)^2$ where $v$ is the forward speed of the HalfCheetah. The control input is the torque applied on each joint, ranging from -1 to 1. The episodes are of length 200.

For robustness evaluation in Section~\ref{sec:exp:evaluation}, we apply an impulsive disturbance torque on each joint every 20 seconds, of which the magnitude ranges from 0.2 to 2.0 and the direction is opposite to the direction of control input. 

\subsection{FetchReach}
We modify the FetchReach in Gym's robotics environment~\cite{brockman2016openai} to a cost version, where the controller is expected to control the manipulator's end effector to reach a random goal position. The cost is designed as $ c= d$, where $d$ is the distance between goal and end-effector. The control input is the torque applied on each joint, ranging from -1 to 1. The episodes are of length 200.

For robustness evaluation in Section~\ref{sec:exp:evaluation}, we apply an impulsive disturbance torque on each joint every 20 seconds, of which the magnitude ranges from 0.2 to 2.0 and the direction is opposite to the direction of control input. 

\subsection{Swimmer}
Swimmer is a modified version based on the environment in the DeepMind control suite~\cite{tassa2018deepmind}. The task is to control a multi-joint snake robot to run at a speed of $1$ $m/s$.
The reward is $r = (v-1)^2$ where $v$ is the forward speed of the Swimmer. The control input is the torque applied on each joint, ranging from -1 to 1. The episodes are of length 250.

For robustness evaluation in Section~\ref{sec:exp:evaluation}, we apply an impulsive disturbance torque on each joint every 20 seconds, of which the magnitude ranges from 0.2 to 1.0 and the direction is opposite to the direction of control input. 

\subsection{Minitaur}
This example is borrowed from the PyBullet environment \cite{coumans2016pybullet}. In Minitaur, the agent controls the Ghost Robotics Minitaur quadruped to run forward at the speed of $1$ $m/s$.
The reward is $r = (v-1)^2$ where $v$ is the forward speed of the robot. The control input is the desired pose of each actuator. The episodes are of length 500.

For robustness evaluation in Section~\ref{sec:exp:evaluation}, we apply an impulsive disturbance on the input channel every 20 seconds, of which the magnitude ranges from 0.2 to 1.0 and the direction is opposite to the direction of control input. 

\section{Synthetic Biology Gene Regulatory Networks}\label{app:repressilator}

Since the gene regulatory networks (GRN) considered here is in nano-scale whose physical property is different from the ones considered in Section~\ref{app:Experiment Setup}. Also, GRN can exhibit interesting oscillatory behavior. We illustrate this example separately in this section. 

\subsection{Mathematical model of GRN }
In this example, we consider a classical dynamical system in systems/synthetic biology, the repressilator, which we use to illustrate the reference tracking task at hand.
The repressilator is a synthetic three-gene regulatory network where the dynamics of mRNAs and proteins follow an oscillatory behavior~\cite{elowitz2000synthetic}.
A discrete-time mathematical description of the repressilator, which includes both
transcription and translation dynamics, is given by the following set of discrete-time equations:
\begin{equation}
\begin{aligned}
x_{1}(t+1) &=x_{1}(t)+dt\cdot \left[-\gamma _{1}x_{1}(t)+\frac{a _{1}}{K_1+x_{6}^{2}(t)} + u_1\right]+\xi_1(t),  \\
x_{2}(t+1) &=x_{2}(t) +dt\cdot\left[-\gamma_{2}x_{2}(t)+\frac{a _{2}}{K_2+x_{4}^{2}(t)} + u_2\right]+\xi_2(t), \\
x_{3}(t+1)&=x_{3}(t)+dt\cdot\left[-\gamma_{3}x_{3}(t)+\frac{a _{3}}{K_3+x_{5}^{2}(t)} + u_3\right]+\xi_3(t),
 \\
x_{4}(t+1) &=x_{4}(t) +dt\cdot\left[-c_{1}x_{4}(t)+\beta _{1}x_{1}(t)\right]+\xi_4(t),  \\
x_{5}(t+1)&=x_{5}(t) +dt\cdot\left[-c_{2}x_{5}(k)+\beta _{2}x_{2}(t)\right]+\xi_5(t),  \\
x_{6}(t+1) &=x_{6}(t) +dt\cdot\left[-c_{3}x_{6}(t)+\beta _{3}x_{3}(t)\right]+\xi_6(t).
\label{oscillator}
\end{aligned}
\end{equation}
Here, $x_{1},x_{2},x_{3}$ (resp. $x_{4},x_{5},x_{6}$) denote the concentrations of the mRNA transcripts (resp. proteins) of genes 1, 2, and 3, respectively.
$\xi_i$, $\forall i$ are i.i.d. uniform noise ranging from $[-\delta,\delta]$, i.e., $\xi_i\sim \mathcal{U}(-\delta,\delta)$. During training, $\delta=0$ and for evaluation $\delta$ is set to $0.5$ and $1$ respectively in Section~\ref{sec:exp:evaluation}.
$a _{1},a
_{2},a _{3}$ denote the maximum promoter strength for their corresponding gene,
$\gamma_{1},\gamma _{2},\gamma _{3}$ denote the mRNA degradation rates, $c_{1},c_{2},c_{3}$ denote the protein degradation rates, $\beta
_{1},\beta _{2},\beta _{3}$ denote the protein production rates, and
$K_{1},K_{2},K_{3}$ are the dissociation constants.
The set of equations in Eq.(\ref{oscillator}) corresponds to a topology where gene $1$ is repressed by gene $2$, gene $2$ is repressed by gene $3$, and gene $3$ is repressed by gene $1$. 
$dt$ is the discretization time step.

\begin{figure}[htb]
    \centering
    \includegraphics[width=0.4\columnwidth]{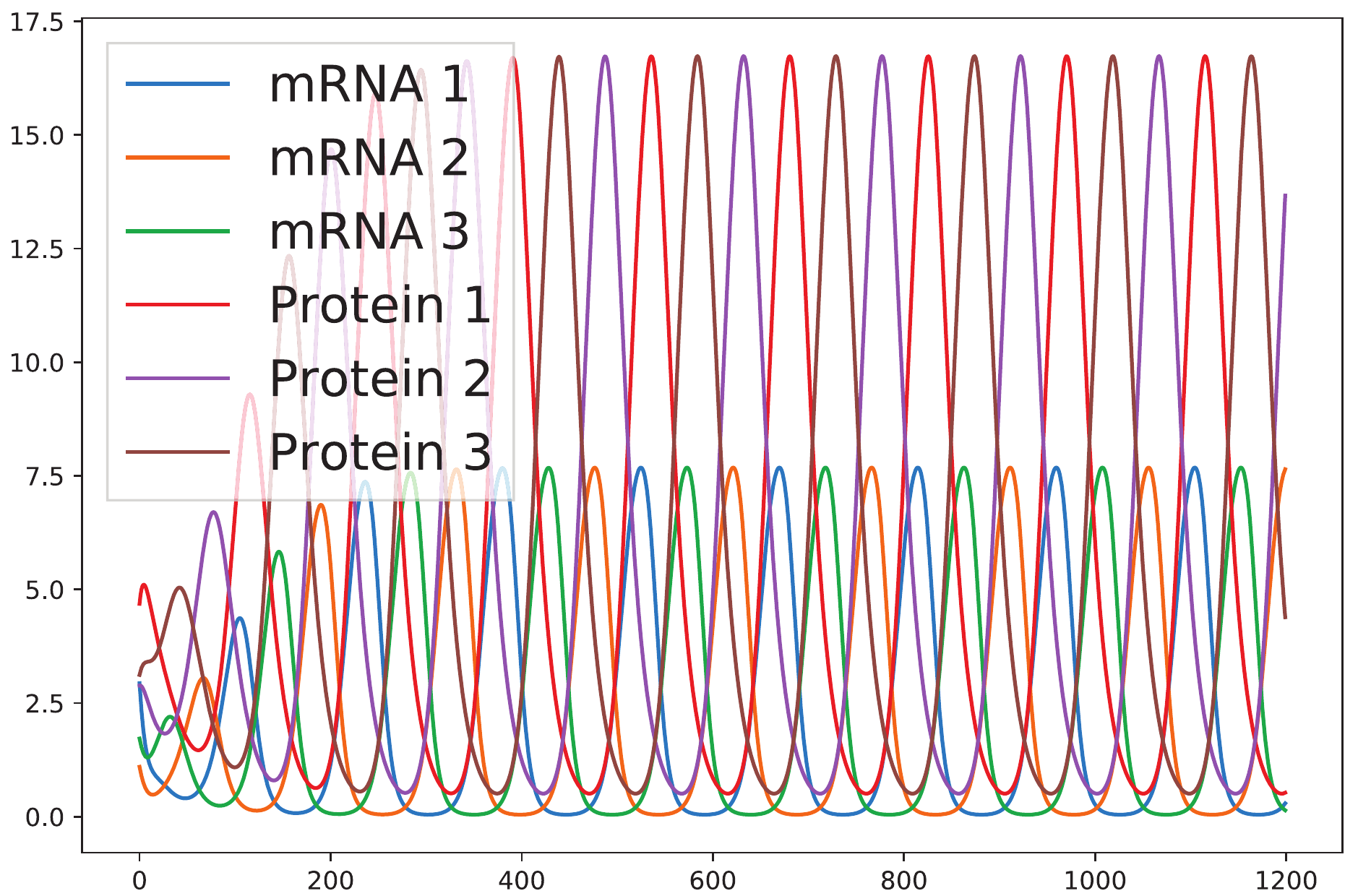}
    \caption{A snapshot of the natural oscillatory behavior of a repressilator system consisting of 3 genes. The oscillations have a period of approximately 150 arbitrary time units. The X-axis denotes time and Y-axis denotes the value/concentration of each state. }
    \label{app:oscillatorsimulation}
\end{figure}

In practice, only the protein concentrations are observed and given as readouts, for instance via fluorescent markers (e.g., green fluorescent protein, GFP or red fluorescent protein, mCherry).
The control scheme $u_i$ will be implemented by light control signals which can induce the expression of genes through the activation of their photo-sensitive promoters. 
To simplify the system dynamics and as it is usually done for the repressilator model~\cite{elowitz2000synthetic}, we consider the corresponding parameters of the mRNA and protein dynamics for different genes to be equal. More background on mathematical modeling and control of synthetic biology gene regulatory networks can be referred to~\cite{strelkowa2010switchable, sootla2013periodic}.
In this example, the parameters are as follows: 
$$\forall i: \,\, K_i=1,  a_{i} = 1.6,  \gamma_{i} = 0.16, \beta_{i} = 0.16, c_{i} = 0.06, dt = 1$$
In Fig\ref{app:oscillatorsimulation}, a single snapshot of the state temporal evolution without $\xi$ is depicted. We uniformly initialized between 0 to 5, i.e., $x_i(0)\sim \mathcal{U}(0,5)$, which is the range we train the policy in Section~\ref{sec:experiment}, persistent oscillatory behavior is also exhibiting similar to the snapshot in Fig~\ref{app:oscillatorsimulation}.

\begin{figure}[htb]   
    \centering
    \subfigure[GRN]{
    \includegraphics[width=0.45\columnwidth]{figure/fig-with-SPPO/return/oscillator-training-return-eps-converted-to.pdf}
    }
    \centering
    \subfigure[CompGRN]{
    \includegraphics[width=0.48\columnwidth]{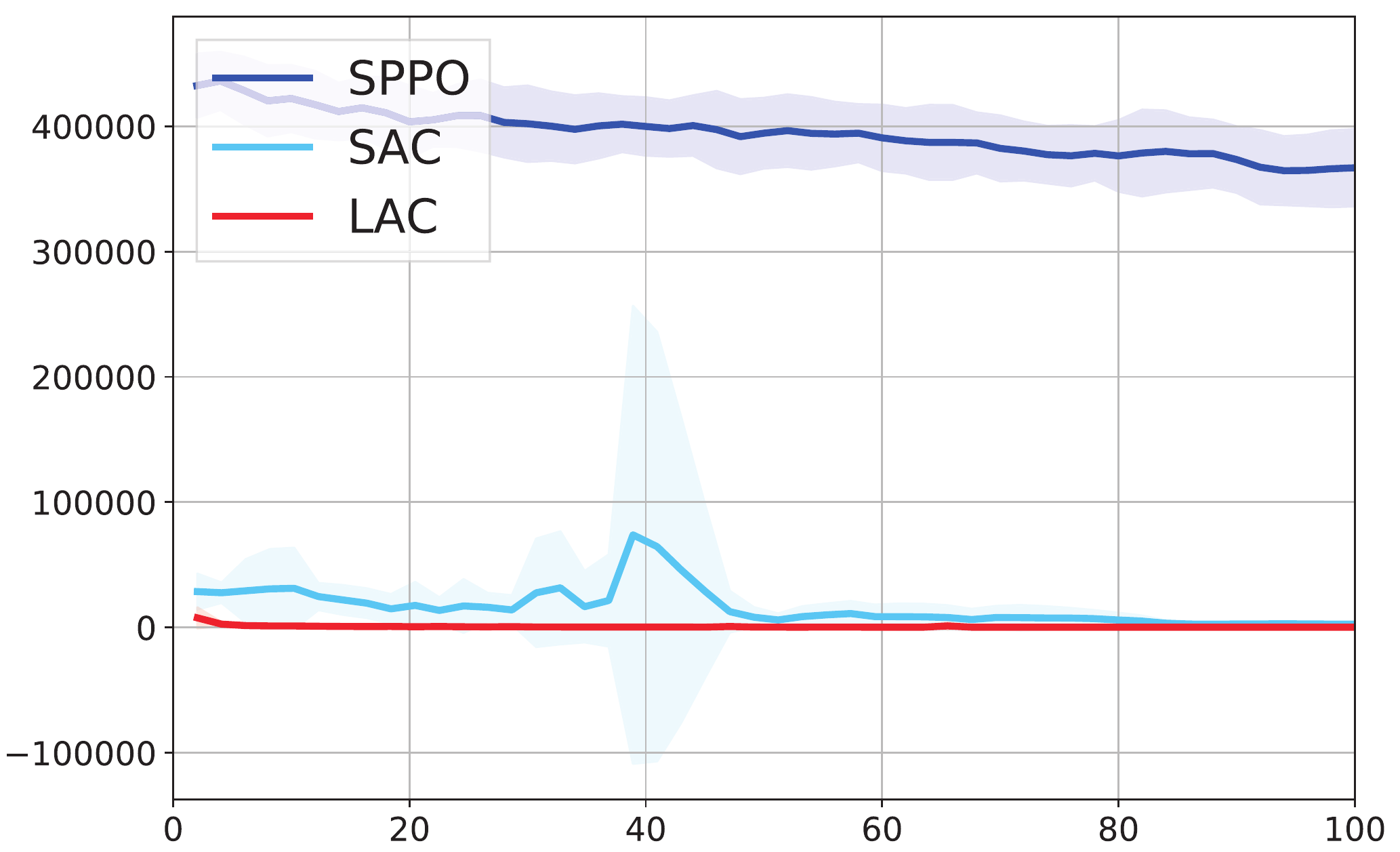}
    \label{GRN-Complicated-supp}
    }
    \caption{Cumulative control performance comparison for synthetic GRN. 
    The Y-axis indicates the total cost during one episode and the X-axis indicates the total time steps in thousand. The shadowed region shows the 1-SD confidence interval over 10 random seeds. Across all trials of training, LAC converges to stabilizing solution with comparable or superior performance compared with SAC and SPPO. Figure.S9 (a) is identical to Figure.1 (d).}
    \label{fig:performance:osc}
\end{figure}

\section{Hyperparameters}\label{app:Hyperparameters}

\begin{table*}[htb]
\caption{Hyperparameters of LAC}\label{table:hyperparameters}
\begin{center}
\begin{tabular}{l|c c c c c c c}

Hyperparameters&CartPole&FetchReach&HalfCheetah&GRN&CompGRN&Swimmer&Minitaur\\\hline
Lyapunov candidate& Sum of cost& Sum of cost& Value& Sum of cost& Sum of cost& Value& Value\\
Time horizon $N$&5&5&$\infty$&$5$&$5$&$\infty$&$\infty$\\
Minibatch size& 256& 256& 256& 256& 256& 256& 256\\
Actor learning rate & 1e-4& 1e-4& 1e-4& 1e-4& 1e-4& 1e-4& 1e-4\\
Critic learning rate & 3e-4& 3e-4& 3e-4& 3e-4& 3e-4& 3e-4& 3e-4\\
Lyapunov learning rate & 3e-4& 3e-4& 3e-4& 3e-4 & 3e-4& 3e-4 & 3e-4\\
Target entropy& -1&-5&-6&-3&-4&-2&-8\\
Soft replacement($\tau$) &0.005&0.005&0.005&0.005&0.005&0.005&0.005\\
Discount($\gamma$)  & NAN & NAN &0.995 & NAN & NAN&0.995&0.995 \\
$\alpha_3$&1.0& 1.0 &1.0&1.0&1.0&1.0&1.0\\
Structure of $f_\phi$ & (64,64,16) &(64,64,16)& (256,256,16) &(256,256,16)& (256,256,16)&(64,64,16)& (256,256,16)\\
\end{tabular}
\end{center}

\end{table*}

For LAC, there are two networks: the policy network and the Lyapunov critic network. For the policy network, we use a fully-connected MLP with two hidden layers of 256 units, outputting the mean and standard deviations of a Gaussian distribution. 
As mentioned in \autoref{sec:algorithm}, it should be noted that the output of the Lyapunov critic network is a square term, which is always non-negative. More specifically, we use a fully-connected MLP with two hidden layers and one output layer with different units as in \autoref{table:hyperparameters}, outputting the feature vector $\phi(s,a)$. The Lyapunov critic function is obtained by $L_c(s,a)=\phi^T(s,a)\phi(s,a)$. All the hidden layers use Relu activation function and we adopt the same invertible squashing function technique as~\cite{haarnoja2018soft} to the output layer of the policy network.

\section{Convergence of Lagrange Multipliers}

\begin{figure}[htb]
    \centering
    \subfigure[Cartpole]{
    \includegraphics[width=0.12\textwidth]{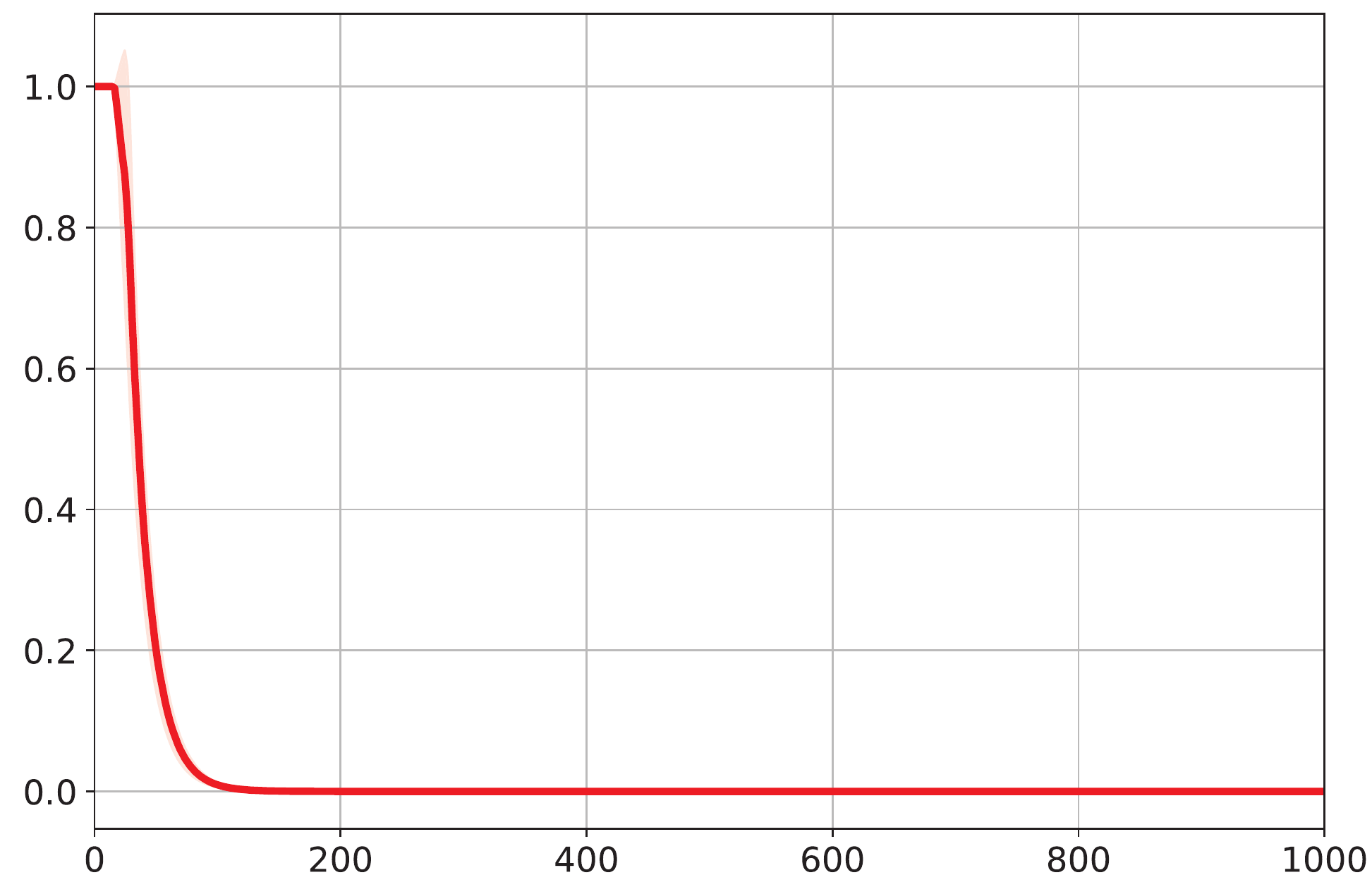}
    }
    \centering
    \subfigure[HalfCheetah]{
    \includegraphics[width=0.12\textwidth]{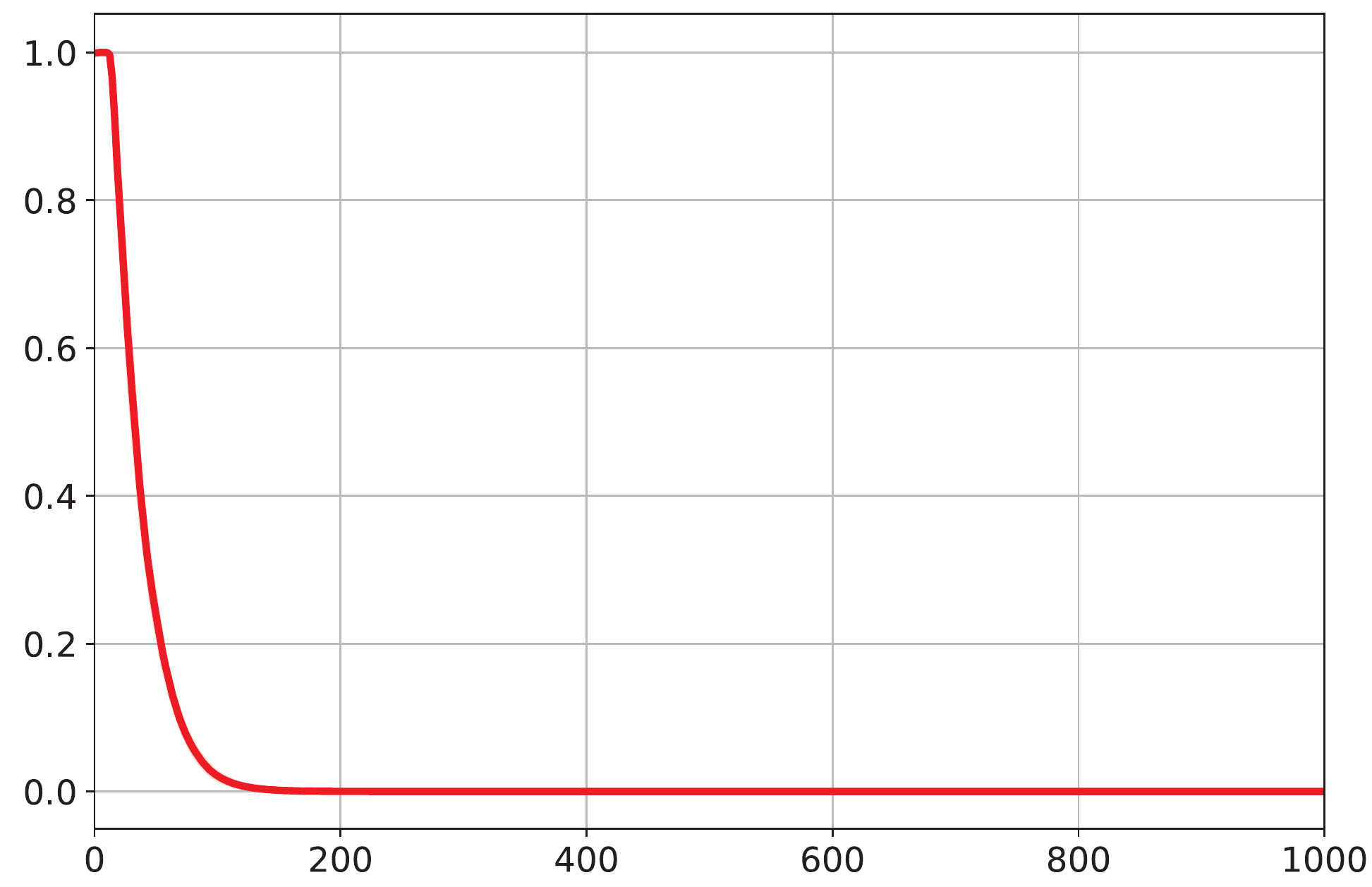}
    }
    \centering
    \subfigure[FetchReach]{
    \includegraphics[width=0.12\textwidth]{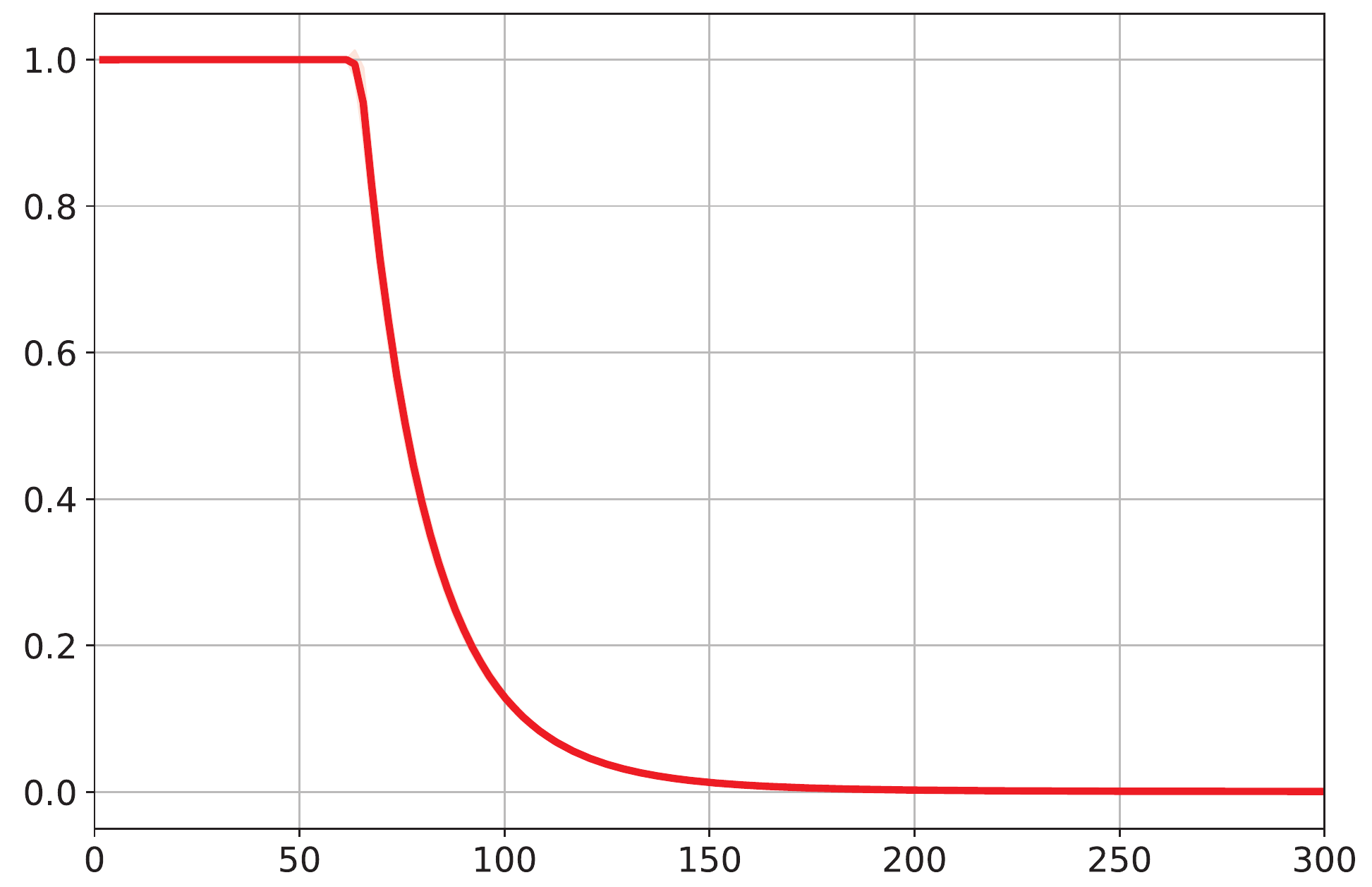}
    }
    \centering
    \subfigure[GRN]{
    \includegraphics[width=0.12\textwidth]{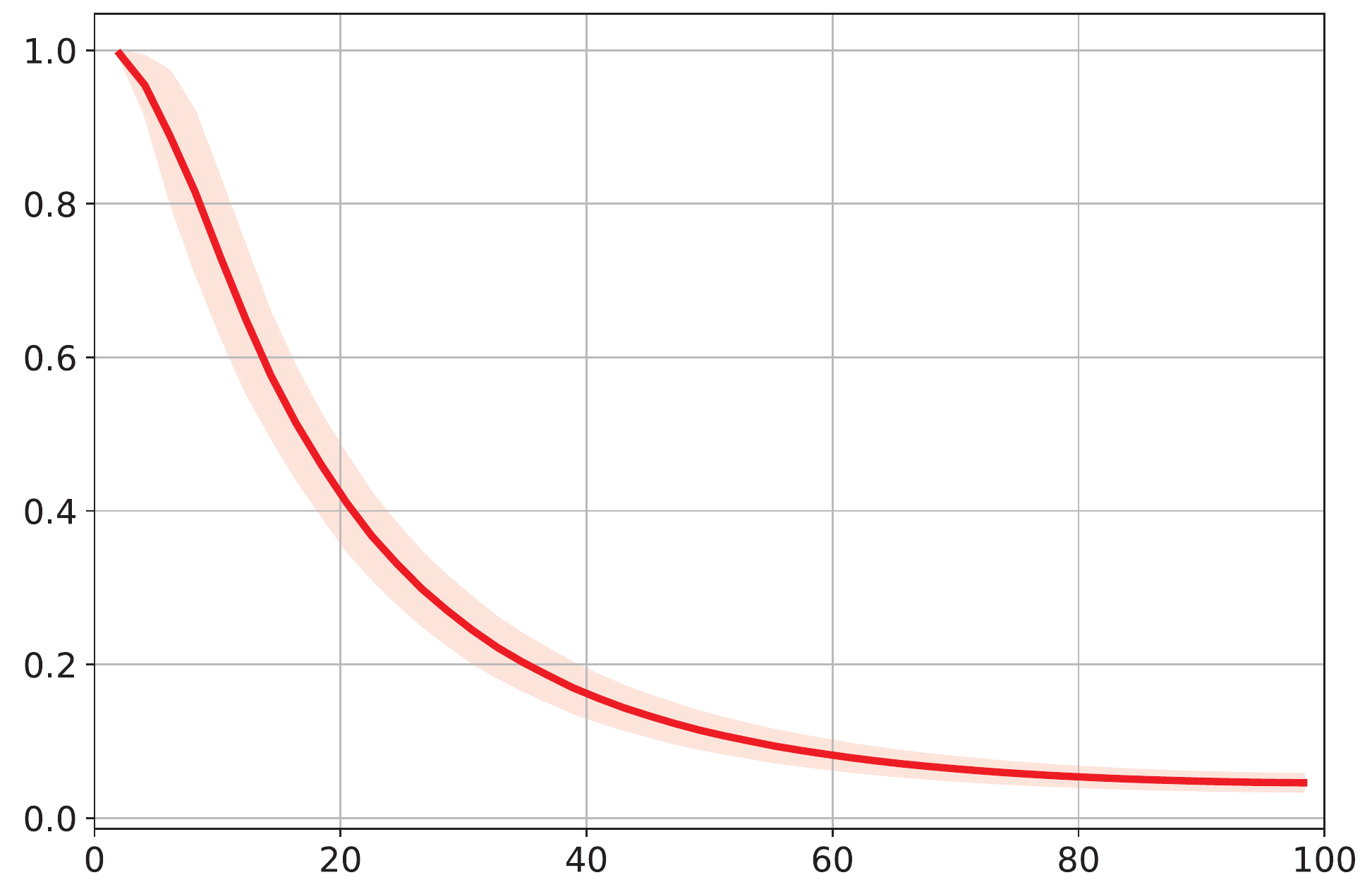}
    }
    \centering
    \subfigure[CompGRN]{
    \includegraphics[width=0.12\textwidth]{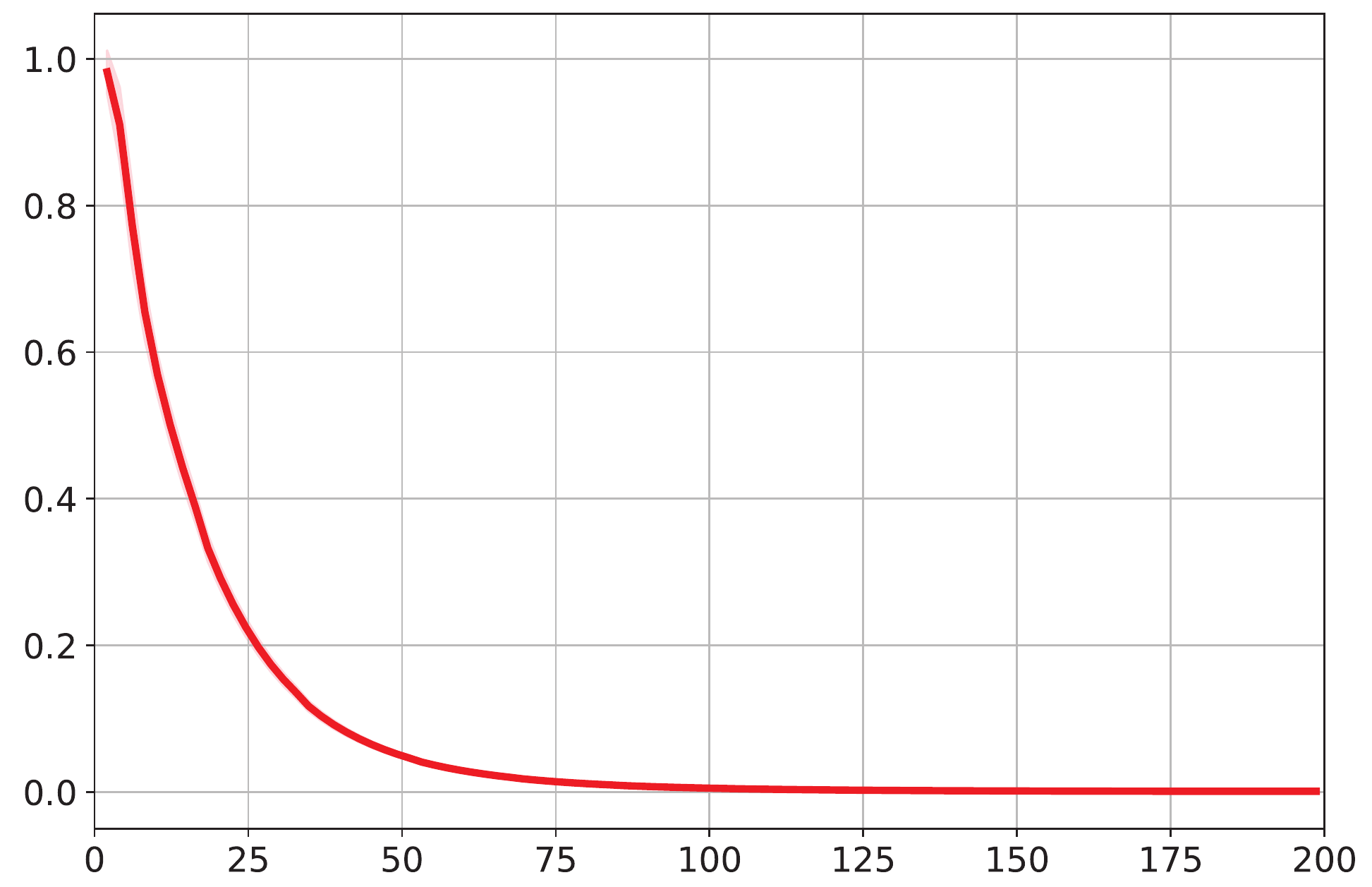}
    }
    \centering
    \subfigure[Swimmer]{
    \includegraphics[width=0.12\textwidth]{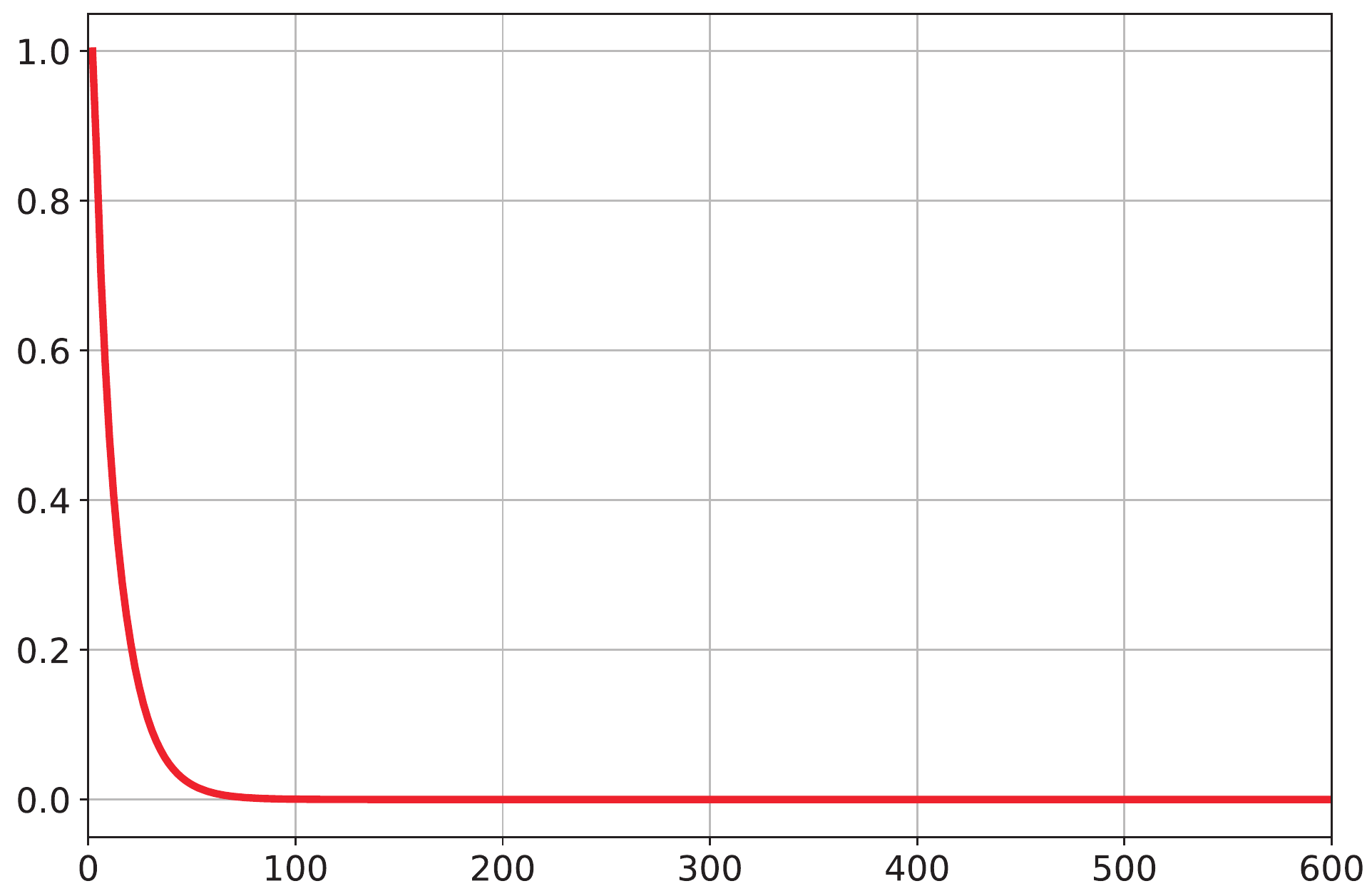}
    }
    \centering
    \subfigure[Minitaur]{
    \includegraphics[width=0.12\textwidth]{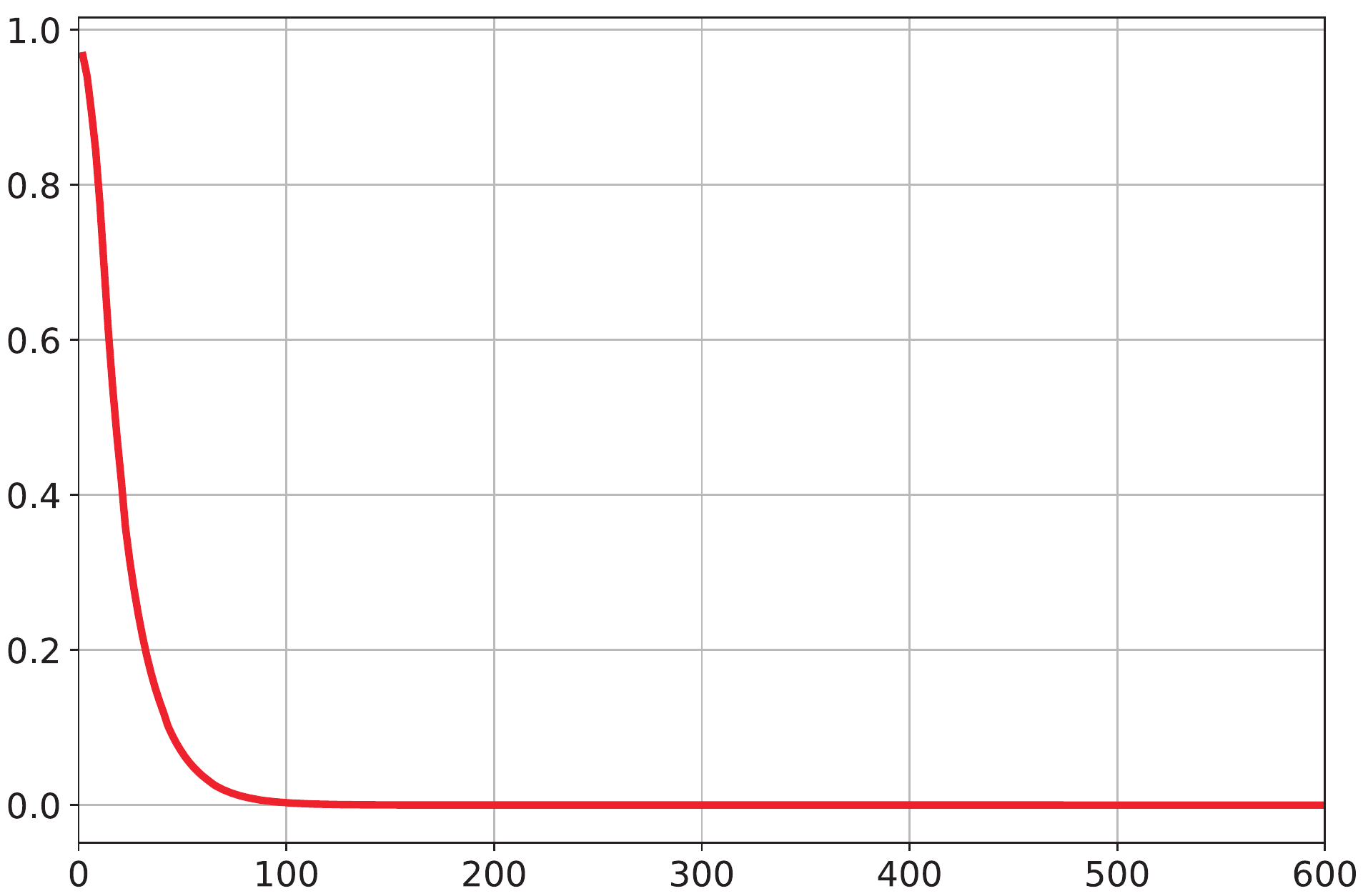}
    }
    \caption{Value of Lagrange multiplier $\lambda$ during the training of LAC policies. The Y-axis indicates the value of $\lambda$ and the X-axis indicates the total time steps in thousand. The shadowed region shows the 1-SD confidence interval over 10 random seeds. The value of $\lambda$ gradually drops and becomes zero at convergence, which implies the satisfaction of stability condition.}
    \label{fig:lagrangian}
\end{figure}

The convergence of LAC and validation of the stability guarantee can also be checked by observing the value of Lagrange multipliers. When (\ref{eq:parameterized lyapunov inequality}) is satisfied, $\lambda$ will continuously decrease until it becomes zero. Thus by checking the value of $\lambda$, the satisfaction of stability condition during training and at convergence could be validated.
In practice, the value of $\lambda$ is clipped under the maximum value $1$ in case that $\lambda$ grows too large due to the violation of stability condition during the early training stage, resulting in the inappropriate step length for the policy update. Clipping is a useful technique to prevent instability of optimization, especially in gradient-based methods, see \cite{schulman2017proximal,bengio2013advances,bello2017neural, wang2015dueling}.
In \autoref{fig:lagrangian}, the value of $\lambda$ during training is demonstrated. Across all training trials in the experiments, $\lambda$ converges to zero eventually, which implies that the stability guarantee is valid. It is also shown that the clipping technique only makes effect in the early training stage in the FetchReach (see (c)) while in other experiments it is not triggered.

\section{Robustness and Generalization Evaluation of SPPO}\label{app:Robustness Evaluation of SPPO}

\begin{figure}[htb]
    \centering
    \subfigure[Cartpole]{
    \includegraphics[width=0.4\columnwidth]{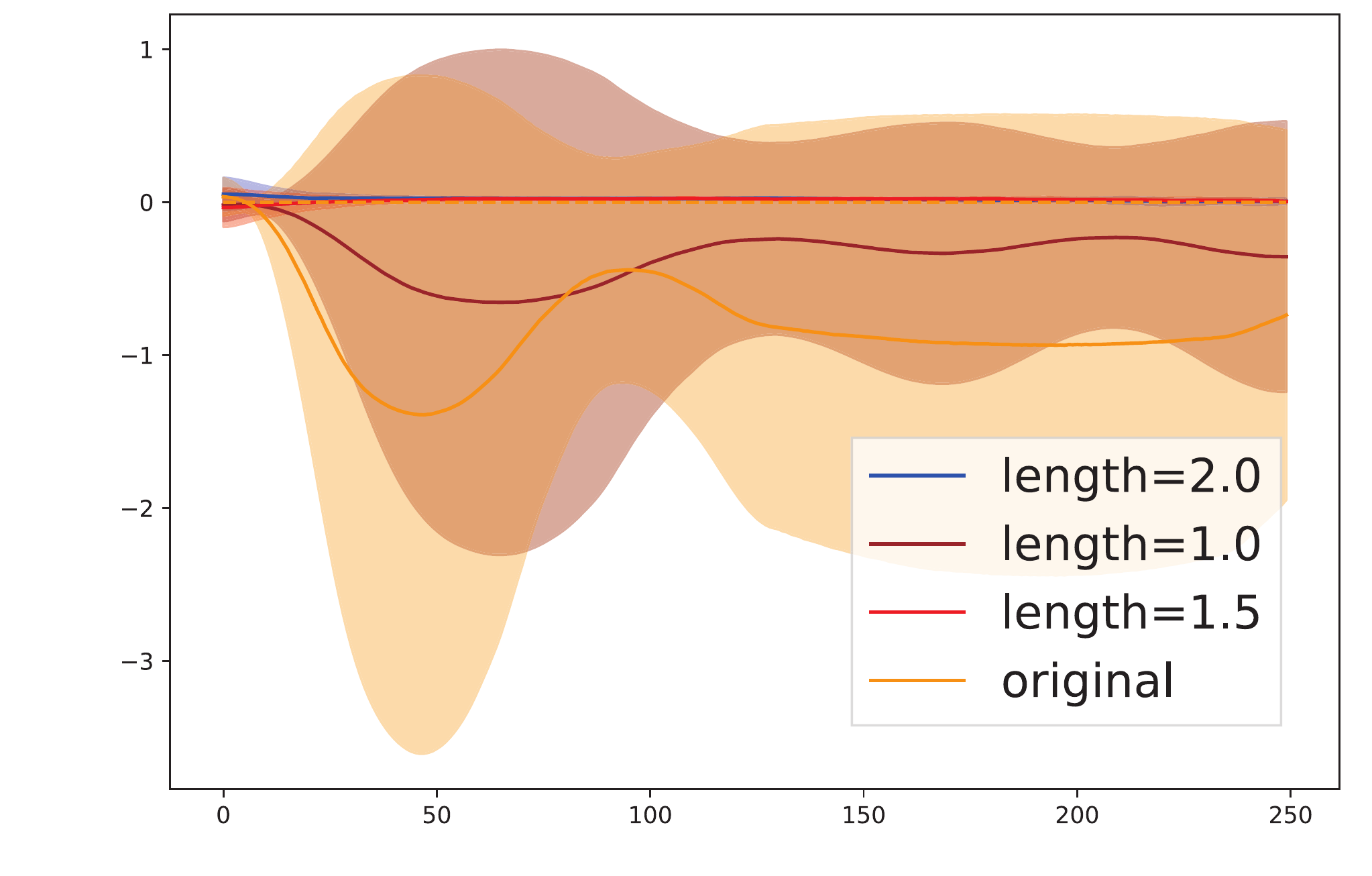}
    }
    \centering
    \subfigure[GRN]{
    \includegraphics[width=0.4\columnwidth]{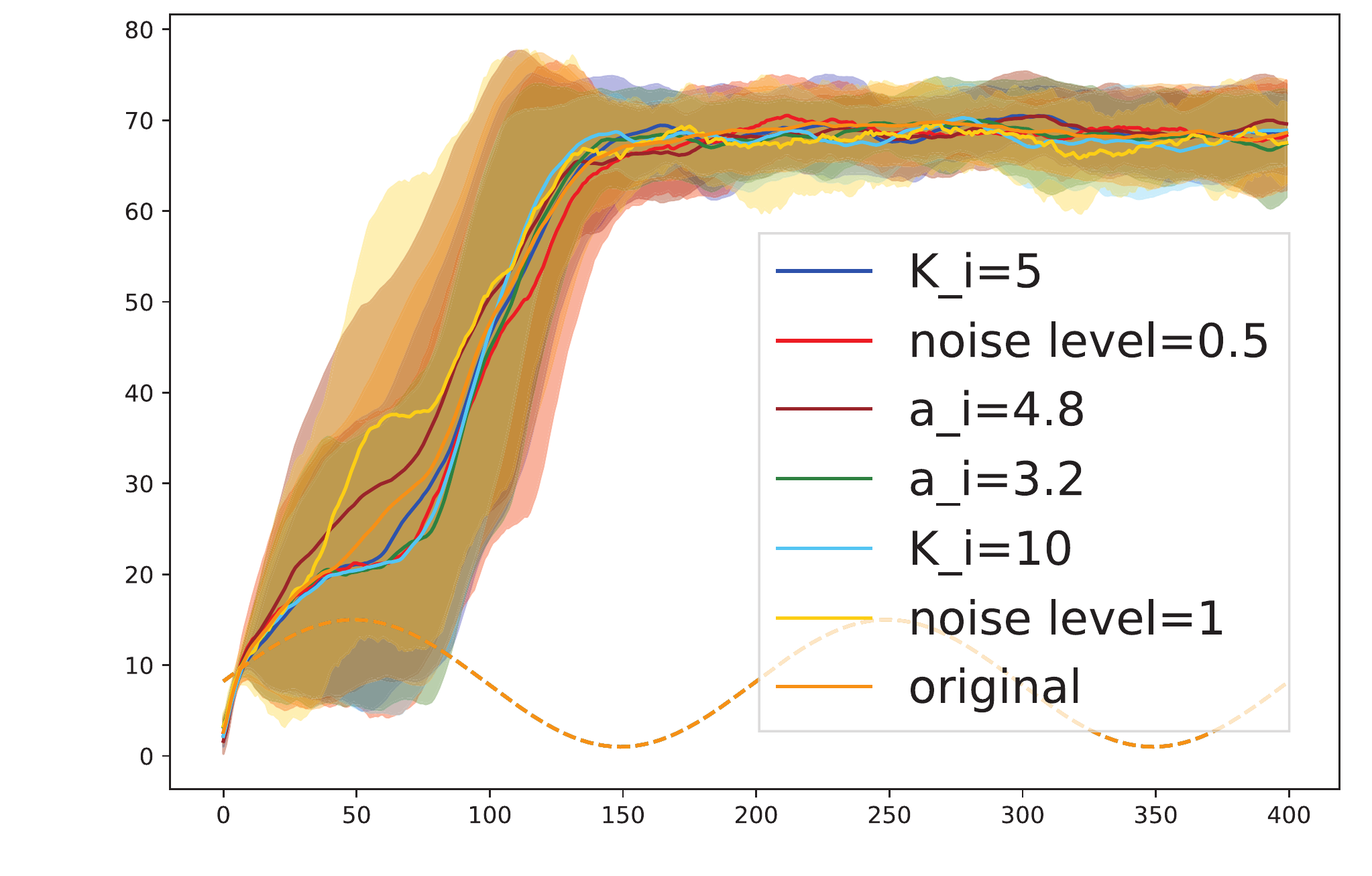}
    }
    \caption{State trajectories over time under policies trained by SPPO and tested in the presence of parametric uncertainties and process noise, for CartPole and  Repressilator. The setting of the uncertainties is the same as in Section~\ref{sec:Robustness to dynamic uncertainty}.
    }
    \label{fig:SPPO-robustness}
\end{figure}

\begin{figure}[htb]
    \centering
    \subfigure[Repressilator]{
    \includegraphics[width=0.4\columnwidth]{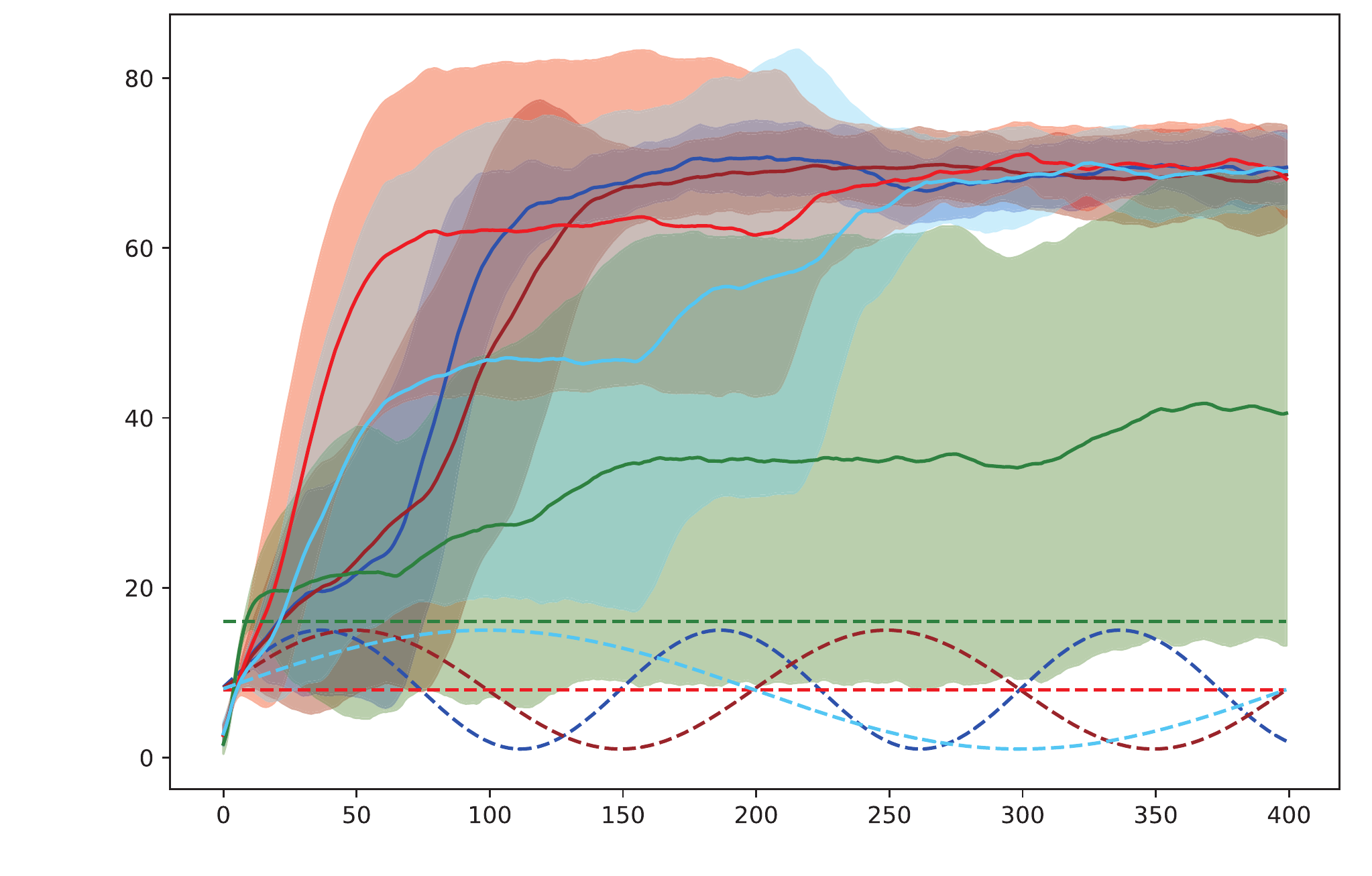}
    }
    \caption{State trajectories under policies trained by SPPO when tracking different reference signals.
    The setting of the uncertainty is the same as in Section~\ref{sec:Generalization over different tracking references}.
    }
    \label{fig:SPPO-generalization}
\end{figure}

In this part, we evaluate the robustness and generalization ability of policies trained by SPPO in the same. First, the robustness of the policies is tested by perturbing the parameters and adding noise in the Cartpole and Repressilator environment, as described in Section~\ref{sec:Robustness to dynamic uncertainty}. Generalization of the policies is evaluated by setting reference signals that are unseen during training. State trajectories of the above experiments are demonstrated in \autoref{fig:SPPO-robustness} and \autoref{fig:SPPO-generalization}, respectively. 
As demonstrated in the figures, the SPPO policies could hardly deal with previously unseen uncertainty or reference signals and failed in all of the Repressilator experiments.

The SPPO algorithm is originally developed for the control tasks with safety constraints, i.e. keeping the expectation of discounted cumulative safety cost below a certain threshold. Though Lyapunov's method is exploited, the approach is not aimed at providing stability guarantee.

\end{document}